\documentclass[runningheads]{llncs}

\usepackage{eccv}

\usepackage{eccvabbrv}

\usepackage{graphicx}
\usepackage{booktabs}

\usepackage[accsupp]{axessibility}  

\usepackage[pagebackref,breaklinks,colorlinks,citecolor=eccvblue]{hyperref}

\usepackage{orcidlink}

\usepackage{pifont}
\usepackage{colortbl}
\usepackage{graphicx}
\usepackage{amsmath}
\usepackage{amssymb}
\usepackage{booktabs}
\usepackage{enumitem}
\usepackage{ulem}
\usepackage{booktabs, multirow} 

\usepackage{xcolor}  
\usepackage{utfsym}  

\usepackage{multirow,algorithm}
\usepackage{colortbl}
\usepackage{enumitem}
\usepackage{amssymb}
\usepackage{makecell}
\usepackage{wrapfig}
\usepackage{bm}
\usepackage{tikz}

\definecolor{barrier}{RGB}{112,128,144}
\definecolor{bicycle}{RGB}{220,20,60}
\definecolor{bus}{RGB}{255, 127, 80}
\definecolor{car}{RGB}{255, 158, 0}
\definecolor{const. veh.}{RGB}{233, 150, 70}
\definecolor{motorcycle}{RGB}{255,61,99}
\definecolor{pedestrian}{RGB}{0,0,230}
\definecolor{traffic cone}{RGB}{47,79,79}
\definecolor{trailer}{RGB}{255,140,0}
\definecolor{truck}{RGB}{255,99,71}
\definecolor{drive. suf.}{RGB}{0,207,191}
\definecolor{other flat}{RGB}{175,0,75}
\definecolor{sidewalk}{RGB}{75,0,75}
\definecolor{terrain}{RGB}{112,180,60}
\definecolor{manmade}{RGB}{222,184,135}
\definecolor{vegetation}{RGB}{0,175,0}
\definecolor{y}{HTML}{00994C}
\definecolor{b}{rgb}{0.31372549, 0.11764706, 0.70588235}

\def\eg{\textit{e.g}\onedot}  
\def\ie{\textit{i.e}\onedot} 

\begin{document}

\title{
\hspace{0.4cm} 
OmniNWM: Omniscient Driving Navigation World Models
}

\titlerunning{OmniNWM}

\makeatletter 
\g@addto@macro\@maketitle{
\begin{tikzpicture}[remember picture,overlay,shift={(current page.north west)}]
\node[anchor=north west, xshift=4.5cm, yshift=-3.7cm]{\scalebox{1}[1]{\includegraphics[width=1.0cm]{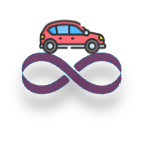}}};
\end{tikzpicture}
}

\footnotesize \author{{Bohan Li$^{1,2}$}\thanks{Equal contribution}, Zhuang Ma$^{3*}$, Dalong Du$^{3*}$, Baorui Peng$^{2}$,\\ Zhujin Liang$^{3}$, Zhenqiang Liu$^{3}$, Xianda Guo$^{6}$, Zheng Zhu$^{5}$,\\
Chao Ma$^{1}$, Yueming Jin$^{4}$,  Xin Jin$^{2}$, Hao Zhao$^{5}$,  Wenjun Zeng$^{2}$\thanks{Corresponding}   
}

\authorrunning{B. Li et al.}




\footnotesize \institute{ $^{1}$Shanghai Jiao Tong University, $^{2}$Ningbo Institute of Digital Twin, Eastern Institute of
Technology, Ningbo, China,
$^{3}$PhiGent, $^{4}$National University of Singapore, $^{5}$Tsinghua University, $^{6}$Wuhan University \\
\vspace{3pt}
\small
\href{https://arlo0o.github.io/OmniNWM/}{https://github.com/Arlo0o/OmniNWM}
}

\maketitle

\vspace{-21pt}
\begin{abstract}
Autonomous driving world models are expected to work effectively across three core dimensions: state, action, and reward.
However, existing methods are typically restricted to fragmented modality modeling, short-horizon drift, and imprecise action control, while lacking intrinsic mechanisms for policy evaluation.
In this paper, we introduce OmniNWM, an Omniscient panoramic Navigation World Model that addresses all three dimensions within a consistent probabilistic framework. For \textit{State}, OmniNWM generates panoramic videos of RGB, semantics, metric depth, and 3D occupancy, ensuring pixel-level alignment across modalities with joint distribution modeling.
To mitigate autoregressive exposure bias, we propose a structured panoramic forcing strategy to stabilize long-horizon generation via stochastic manifold thickening.
For \textit{Action}, we introduce canonical geometric action encoding with normalized panoramic Plücker ray-maps. This representation decouples motion dynamics from sensor intrinsics, enabling precise, zero-shot trajectory control across heterogeneous datasets and camera configurations.
For \textit{Reward}, we derive intrinsic occupancy-grounded dense rewards directly from generated 3D volumes, establishing a reliable closed-loop simulation cycle for evaluating diverse planning agents.
Extensive experiments demonstrate that OmniNWM achieves SOTA performance in generation fidelity and control precision, with remarkable zero-shot robustness to novel scenes on nuPlan and in-house datasets with distinct camera rigs.\looseness=-1
\end{abstract}

\begin{table}[!t]
\centering
\caption{\textbf{Comparison of world model capabilities.} OmniNWM unifies the three core dimensions (state, action, and reward) for autonomous driving. `R', `S', `D', and `O' denote RGB, semantic, depth, and occupancy modalities.  
`$\sim$' represents limited cross-rig transferability due to overfitting with extrinsic-sensitive conditioning.}
\vspace{-9pt}
\scriptsize
\setlength{\tabcolsep}{9.0pt}
\resizebox{1.0\linewidth}{!}{
\begin{tabular}{lcccccc}
\toprule 
\multirow{3}{*}{Method} & \multicolumn{2}{c}{\textbf{State}} & \multicolumn{2}{c}{\textbf{Action}} & \multicolumn{2}{c}{\textbf{Reward}} \\  
\cmidrule(lr){2-3} \cmidrule(lr){4-5} \cmidrule(lr){6-7}
 & Modalities & Length & Control Signal & \makecell[c]{Cross-Rig\\Transfer} & Source & \makecell[c]{Closed\\Loop} \\  
\midrule 
DriveGAN~\cite{kim2021drivegan} & R & 6 & Waypoint & $\sim$ & - & \textcolor{red}{\usym{2717}} \\ 
Drivedreamer~\cite{wang2023drivedreamer} & R & 32 & Vel, Angle & $\sim$ & - & \textcolor{red}{\usym{2717}} \\ 
Drivedreamer-2~\cite{zhao2024drivedreamer2} & R & 32 & Vel, Angle & $\sim$ & - & \textcolor{red}{\usym{2717}} \\ 
Vista~\cite{gao2025vista}& R & 24 & Waypoint & $\sim$ & - & \textcolor{red}{\usym{2717}} \\
DrivingGPT~\cite{chen2024drivinggpt} & R & 60 & Waypoint & $\sim$ & - & \textcolor{red}{\usym{2717}} \\ 
MagicDrive~\cite{gao2024magicdrive} & R & 60 & Waypoint, Poses & $\sim$ & - & \textcolor{red}{\usym{2717}} \\ 

\midrule
WoVoGen~\cite{lu2023wovogen}& R,O & 6 & 3D Volume & \textcolor{red}{\usym{2717}} & - & \textcolor{red}{\usym{2717}} \\ 
OccScene~\cite{li2025occscene} & R,O & 8 & 3D Volume & \textcolor{red}{\usym{2717}} & - & \textcolor{red}{\usym{2717}} \\ 
\midrule
Drive-WM~\cite{wang2023drivewm} & R & 8 & Waypoint & $\sim$ & \makecell[c]{{External}\\Image-based Model} & \textcolor{red}{\usym{2717}} \\ 
\midrule
\rowcolor{gray!10} {OmniNWM (Ours)} & {R,S,D,O} & {321} & \makecell[c]{{Normalized}\\Panoramic {Ray-map}} & \textcolor{green}{\usym{2713}} & \makecell[c]{{Intrinsic}\\{3D Semantic Occupancy}} & \textcolor{green}{\usym{2713}} \\ 
\bottomrule
\end{tabular}
}
\vspace{-0pt}
\vspace{-23pt}
\label{tab_down1}
\end{table}

\vspace{-27pt}
\section{Introduction}
\vspace{-6pt}

Recent advancements in world models have demonstrated significant potential for autonomous driving, facilitating high-fidelity simulation of complex environments and controllable navigation for autonomous agents~\cite{wang2023drivedreamer,li2025uniscene,wang2023drivewm}.
Generally, an ideal world model should approximate the joint multimodal posterior of the real-world environment: predicting future states, evaluating actions, and assigning rewards within a unified probabilistic framework~\cite{gao2023magicdrive,gao2025vista,li2026articulated,wang2023drivewm,li2025uniscene,mao2024dreamdrive,li2026scaling}.

However, achieving this unification remains an open challenge due to the complexity of real-world 3D environments~\cite{guo2025genesis,mousakhan2025orbis,yang2025x,zhou2025hermes,xiao2025robotron,yang2024drivearena,lu2025infinicube,li2025uniscene}. As summarized in Tab.~\ref{tab_down1}, existing methods face three fundamental bottlenecks:
\noindent{1) Long-term Joint Consistency in State:} Current models predominantly rely on single-modality RGB videos and typically treat multi-modal sensor simulation as independent conditional generation tasks, which could lead to a {conditional independence fallacy}~\cite{zheng2023occworld,gao2023magicdrive,wang2023drivedreamer}. Furthermore, existing approaches suffer from exposure bias, causing rapid degradation in long-term temporal coherence for extended rollouts~\cite{li2025occscene,gao2024magicdrive,li2025uniscene,lingbotworld}.
\noindent{2) Geometric Covariate Shift in Action:} Precise camera control is hindered by the entanglement of {motion dynamics} with {sensor rig geometry}. Existing sparse representations (\eg, waypoints, raw camera poses) fail to generalize because the latent space overfits to specific extrinsic calibrations, prohibiting precise controllability and zero-shot transfer across novel datasets and trajectory actions~\cite{wang2023drivedreamer,zhao2024drivedreamer2,gao2024magicdrive,wang2023drivewm,guo2025dist}.
\noindent{3) Lack of Intrinsic Rewards:} A valid world model must provide physically grounded reward signals. 
Although few recent studies propose image-based rewards, they rely on external, black-box reward models that suffer from distribution shift, failing to effectively close the loop between generation and planning for autonomous driving~\cite{wang2023drivewm}.\looseness=-1

\begin{figure*}[!t]
 \vspace{-10pt}
  \setlength{\linewidth}{\textwidth}
  \setlength{\hsize}{\textwidth}
  \centering
    \includegraphics[width=0.92\linewidth]{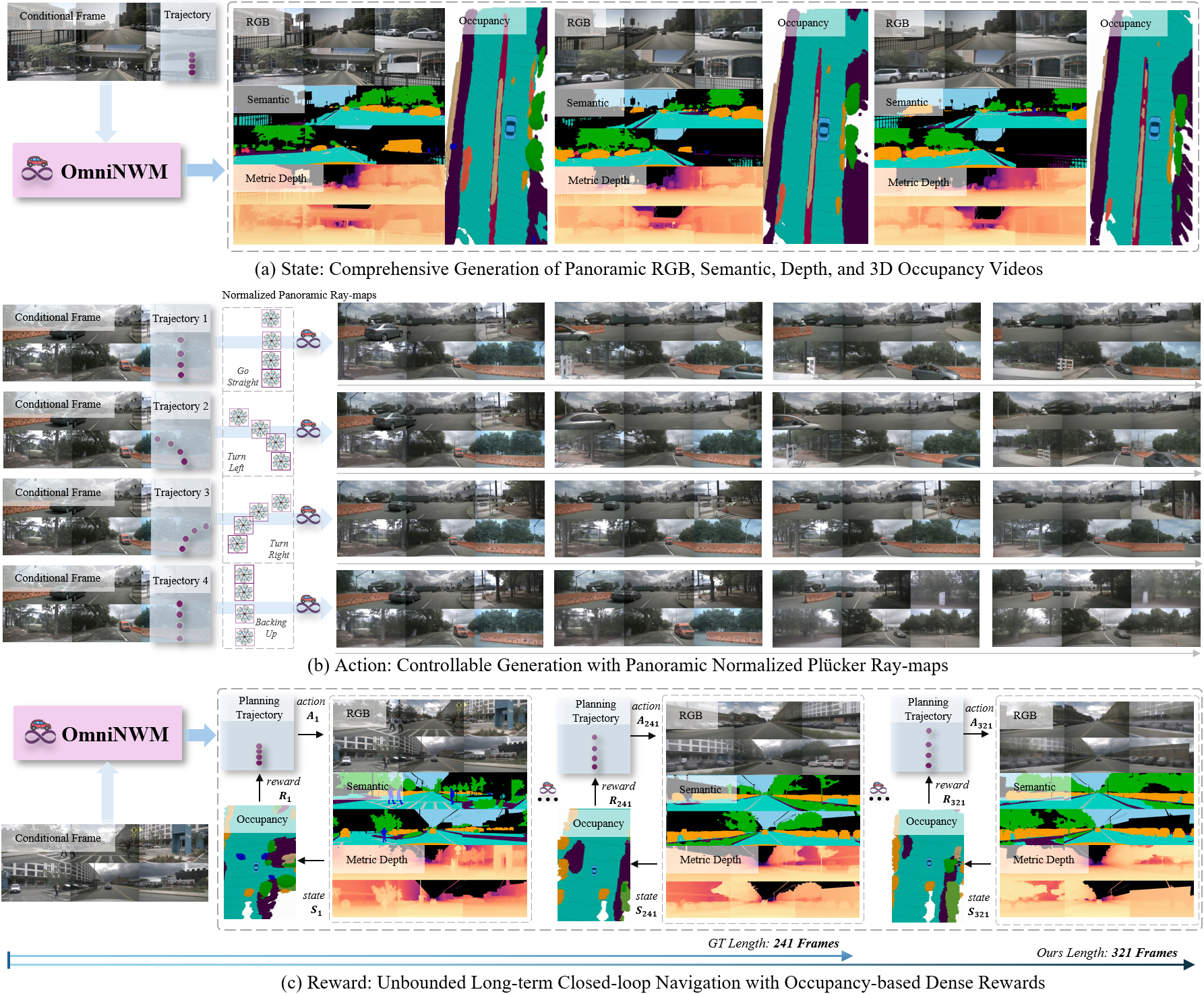}
     \vspace{-9pt}
  \caption{\textbf{Versatile capabilities of OmniNWM.} (a) {Comprehensive State.} OmniNWM generates pixel-aligned panoramic RGB, semantics, metric depth, and 3D occupancy.
  (b) {Canonical Action Control.} We introduce {normalized panoramic Plücker ray-maps} to enable precise control that generalizes across novel trajectories (\eg, reversing, turning). 
  (c) {Intrinsic Reward.} OmniNWM enables long-term navigation (beyond GT length) through a closed-loop pipeline: the future trajectory produced from the planning agent (\eg, OmniNWM-VLA) guides the multi-modal generation, while dense rewards are natively derived from the generated 3D semantic occupancy. 
  }
 \vspace{-20pt}
\label{fig_teaser1}
\end{figure*}
To address these challenges, we propose OmniNWM, an {Omniscient panoramic Navigation World Model} that unifies state, action, and reward within a joint probabilistic framework.
First, to ensure {state consistency}, OmniNWM optimizes the joint distribution of
panoramic RGB, semantics, and metric depth from a shared latent manifold (Fig.~\ref{fig_teaser1} (a)). This joint modeling facilitates pixel-level alignment, thereby lifting the 2D observations into consistent 3D semantic occupancy. 
To stabilize long-term generation, we employ a {structured panoramic forcing} strategy to mitigate autoregressive drift via stochastic manifold thickening. 
Second, to address geometric covariate shift, we introduce a {normalized panoramic Plücker ray-map} encoding scheme (Fig.~\ref{fig_teaser1} (b)). By projecting input trajectories into a canonical geometric space, we decouple motion control from rig calibration, enabling precise trajectory control that generalizes zero-shot across novel datasets and trajectories.
Third, we establish a {closed-loop navigation pipeline}. The generated 3D semantic occupancy serves as an intrinsic utility function, providing dense rewards for the planning agent (\eg, OmniNWM-VLA), which in turn conditions future generation (Fig.~\ref{fig_teaser1} (c)).

Our main contributions are summarized as follows: 
(1) We propose OmniNWM, a unified framework that models the State-Action-Reward Triad. By jointly optimizing panoramic RGB, semantics, and depth, we resolve the {modality drift} inherent in modular architectures and ensure the generated 3D semantic occupancy (and thus the intrinsic reward) is consistent with visual observations.
(2) A Canonical Geometric Action Encoding scheme is introduced with the Normalized Panoramic Ray-map representation, which enables precise action control and zero-shot generalization across distinct camera rigs and unseen scenes.
(3) A Structured Panoramic Forcing strategy is developed to explicitly mitigate autoregressive drift and error accumulation, enabling stable and robust long-term forecasting beyond Ground-Truth (GT) horizons.
(4) We demonstrate a fully Closed-loop Simulation Cycle wherein intrinsic occupancy-grounded dense rewards evaluate the planning agents (\eg, OmniNWM-VLA), which in turn reason and plan future trajectories.\looseness=-1

Experiments demonstrate OmniNWM achieves state-of-the-art (SOTA) generation quality and control precision.
It also exhibits robust zero-shot generalization across unseen datasets (\eg, nuPlan, in-house), varying camera rigs (\eg, 3 and 6 views), and novel trajectories (\eg, reversing), alongside long-term generation beyond GT horizons. Code and videos are provided in the supplementary.\looseness=-1

\vspace{-6pt}
\section{Related Work}
\vspace{-6pt}
\noindent\textbf{World Models for Autonomous Driving.} Recent progress in driving world models has advanced visual realism and controllability~\cite{ye2025bevdiffuser,gao2025vista,zhou2024hugsim,li2024hierarchical,gao2024magicdrive,li2024bridging,wang2023drivewm,chen2025unimlvg,yanginstadrive,xu2025ad,ge2025unraveling,yang2025drivingview,li2026hierarchical,song2025coda,zhao2025recondreamer++,li2025scaling,mao2024dreamdrive,li2026hierarchical,jiang2025rayzer,jin2024lvsm,li2025rig3r,yang2025drivearena,schneider2025neural,li2025uniscene,wang2024stag,kong20253d,hu2025vision,liang2025worldlens,yang2025resim,ni2025recondreamer,liu2025wanderland,liang2026UniFuture}. Early efforts like DriveDreamer~\cite{wang2023drivedreamer} and Drive-WM~\cite{wang2023drivewm} leverage multi-stage diffusion pipelines to model future states. UniFuture~\cite{liang2026UniFuture} proposes to integrate future scene generation of appearance and depth within a framework.  MagicDrive~\cite{gao2023magicdrive} introduces cross-view attention for street-view synthesis, while Vista~\cite{gao2025vista} proposes latent replacement strategies. 
To address 3D structural constraints, recent approaches have integrated occupancy grids~\cite{li2025uniscene}, Gaussian Splatting~\cite{zhou2024hugsim}, and egocentric priors~\cite{hassan2025gem}.
However, existing models typically treat modalities in isolation~\cite{wang2023drivedreamer,gao2023magicdrive,gao2025vista}, and hinder closed-loop evaluation without integrated reward modeling~\cite{zhao2024drivedreamer2,gao2024magicdrive,li2025uniscene}. 
OmniNWM addresses these limitations by unifying long-horizon multi-modal generation and intrinsic dense rewards within a single probabilistic framework.\looseness=-1

\noindent\textbf{Camera-controlled Video Generation.}
Precise camera trajectory control is pivotal for consistent 3D scene synthesis~\cite{he2024cameractrl,li2025cameras,he2025cameractrl,wang2024motionctrl,kuang2024collaborative,bai2025recammaster,yang2024direct,hou2024training,zhang2024cameras,chen2025geodrive}. CameraCtrl~\cite{he2024cameractrl} introduces a plug-and-play module to parameterize camera trajectories. 
MotionCtrl~\cite{wang2024motionctrl} enables flexible motion control with the inherent properties of cameras, while CamCo~\cite{xu2024camco} and VD3D~\cite{bahmani2024vd3d} incorporate epipolar constraints and Plücker embeddings to enforce geometric structure. ReCamMaster~\cite{bai2025recammaster} focuses on re-rendering existing videos from novel camera paths.  
However, these methods are largely restricted to short-horizon, monocular sequences, tending to overfit on specific rig geometries and preventing cross-dataset transfer~\cite{he2024cameractrl,he2025cameractrl}. OmniNWM overcomes these limitations by introducing normalized Plücker ray-maps, enabling precise, spatially consistent control over long-term panoramic sequences and facilitating zero-shot generalization by decoupling motion from calibration.\looseness=-1

\vspace{-12pt}
\section{Methodology}
\vspace{-6pt}
 
\subsection{Comprehensive Generation within OmniNWM}\label{sec_comprehensive}
\vspace{-3pt}

\begin{figure*}[!t]
 \vspace{-10pt}
    \centering    
\includegraphics[width=1.\linewidth]{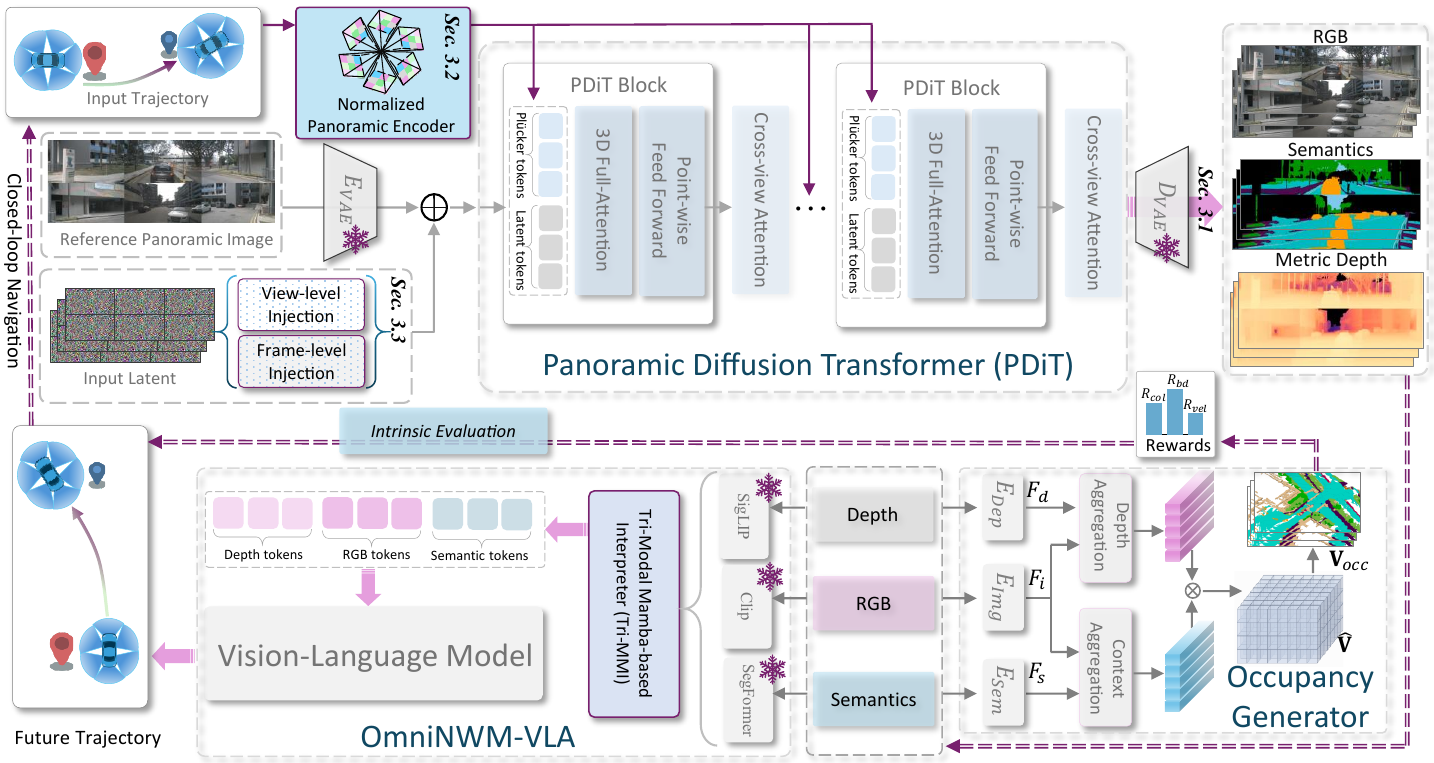}
    \vspace{-15pt}
\caption{\textbf{Architecture overview of OmniNWM.} 
Our framework functions as a unified probabilistic simulation loop. {State}: it jointly forecasts panoramic RGB, semantic, and metric depth videos via a PDiT, facilitating cross-modal consistency for 3D occupancy generation (Sec.~\ref{sec_comprehensive}). {Reward \& Plan}: the generated states provide intrinsic dense rewards and serve as the multi-modal context for the integrated {OmniNWM-VLA} to reason and plan future trajectories (Sec.~\ref{sec_trajectory}). {Action}: these planning trajectories are encoded into {canonical normalized panoramic Plücker ray-maps} (Sec.~\ref{sec_raymap}), closing the loop by guiding the next stage of generation. All components are stabilized by a {structured panoramic forcing} strategy to ensure long-horizon robustness (Sec.~\ref{sec_forcing}).
}
    \label{fig_overall}
    \vspace{-15pt}
\end{figure*}

\noindent\textbf{Pixel-aligned Panoramic Diffusion Transformer.}\label{sec_pixel} Rather than minimizing separate reconstruction losses~\cite{li2025occscene,gao2024magicdrive,li2025uniscene,lingbotworld}, we jointly optimize panoramic multimodal states (Fig.~\ref{fig_overall}). Specifically, our Panoramic Diffusion Transformer (PDiT) avoids the inter-modal misalignment of modular approaches by projecting multi-modalities into a shared latent manifold to enforce holistic scene consistency.\looseness=-1

\noindent\textit{Joint Latent Encoding.} We leverage a pretrained 3D VAE~\cite{kong2024hunyuanvideo} to compress the high-dimensional input space into compact spatiotemporal latents. To preserve semantic topology during this continuous projection, semantic maps are colorized via a fixed palette before encoding and discretized via nearest-neighbor matching after decoding. The latents for RGB, colorized semantics, and metric depth are concatenated channel-wise, forming a unified latent $\mathbf{z}^\texttt{joint} \in \mathbb{R}^{C_\texttt{total} \times T \times H \times W}$.

\noindent\textit{Unified Denoising Dynamics.} The PDiT approximates conditional reverse diffusion $p_\theta(\mathbf{z}_{t-1}^\texttt{joint} | \mathbf{z}_t^\texttt{joint} , \mathcal{C})$ on this joint latent.  
This shared optimization trajectory facilitates pixel-level alignment: as the depth and semantic gradients backpropagate to the joint latent variables as RGB gradients, the outputs are intrinsically synchronized, providing a robust foundation for subsequent 3D occupancy lifting.\looseness=-1

\noindent\textbf{Geometric Lifting for 3D Occupancy.}\label{sec_occupancy} Unlike previous heavy volumetric diffusion~\cite{zheng2023occworld,li2025uniscene}, we propose a lightweight geometric mapping that lifts 2D observations into a 3D grid $\Phi: \{\mathbf{x}_\texttt{rgb}, \mathbf{x}_\texttt{depth}, \mathbf{x}_\texttt{sem}\} \to \mathbf{V}_\texttt{occ}$ (Fig.~\ref{fig_overall}). An EfficientNet-B7~\cite{tan2019efficientnet} U-Net extracts image features $F_{i}$, which are aggregated with depth $F_{d}$ and semantic features $F_{s}$ via 3D convolutions. Following~\cite{li2026hierarchical,li2024bridging}, their outer product computes voxel volume $\mathbf{\hat{V}}$. An upsampling and softmax head then produces the final occupancy $\mathbf{V}_\texttt{occ}$. This efficient process ensures the occupancy remains grounded in visual observations for precise physical evaluation of driving scenes.\looseness=-1

\noindent\textbf{Intrinsic Occupancy-Grounded Rewards.}\label{sec_reward} A valid world model needs to provide effective signals for closed-loop policy evaluation. Moving beyond learned reward models, which can suffer from distribution shift~\cite{wang2023drivewm}, we introduce an intrinsic physical grounded utility function. By querying the generated 3D occupancy volume, we derive a dense potential field that evaluates the safety and compliance of ego trajectories. The reward $\widehat{\mathbf{R}}$ is formulated with physical constraints computed on the generated representation:\looseness=-1
\vspace{-6pt}
\begin{equation}
\widehat{\mathbf{R}}= 1 + (R_{\texttt{col}} + R_{\texttt{bd}} 
+ R_{\texttt{vel}})/N_\texttt{reward},
\end{equation}
where each component represents a dense penalty derived from the voxel grid:
\begin{itemize}
\vspace{-5pt}
\item \textit{Collision Penalty:} Evaluates intersections between the ego-volume and occupied voxels labeled as obstacles (\eg, `car', `barrier'). It scales with velocity $v$ to reflect kinetic energy risk: $R_{\texttt{col}} = - \alpha_{\texttt{col}} \cdot \mathbb{I}_{\texttt{col}} \cdot |v|, \ \text{where} \ \alpha_{\texttt{col}}=0.5$.
\item \textit{Drivable Area Constraint:} Penalizes deviations from `drivable surface' class, enforcing driving compliance: $R_{\texttt{bd}} = - \alpha_{\texttt{bd}} \cdot \mathbb{I}_{\texttt{non-drivable}}, \ \text{where} \ \alpha_{\texttt{bd}}=0.3$.
\item \textit{Traffic Flow Efficiency:} Encourages maintaining target traffic velocity for traffic efficiency:  
$R_{\texttt{vel}} = -\alpha_{\texttt{vel}} \cdot  \texttt{tanh}(|v - v_{\texttt{target}}|) \cdot \mathbb{I}_{v}, \ \text{where} \ \alpha_{\texttt{vel}}=0.2$.
\vspace{-5pt}
\end{itemize}
This occupancy-grounded formulation provides physically consistent signals, enabling closed-loop evaluation of planning agents within the generated world (Fig.~\ref{fig_reward}). More evaluations on reward hyperparameters are in the supplementary.\looseness=-1

\begin{figure}[!t]
     \vspace{-0pt}
    \centering
    \includegraphics[width=1.0\linewidth]{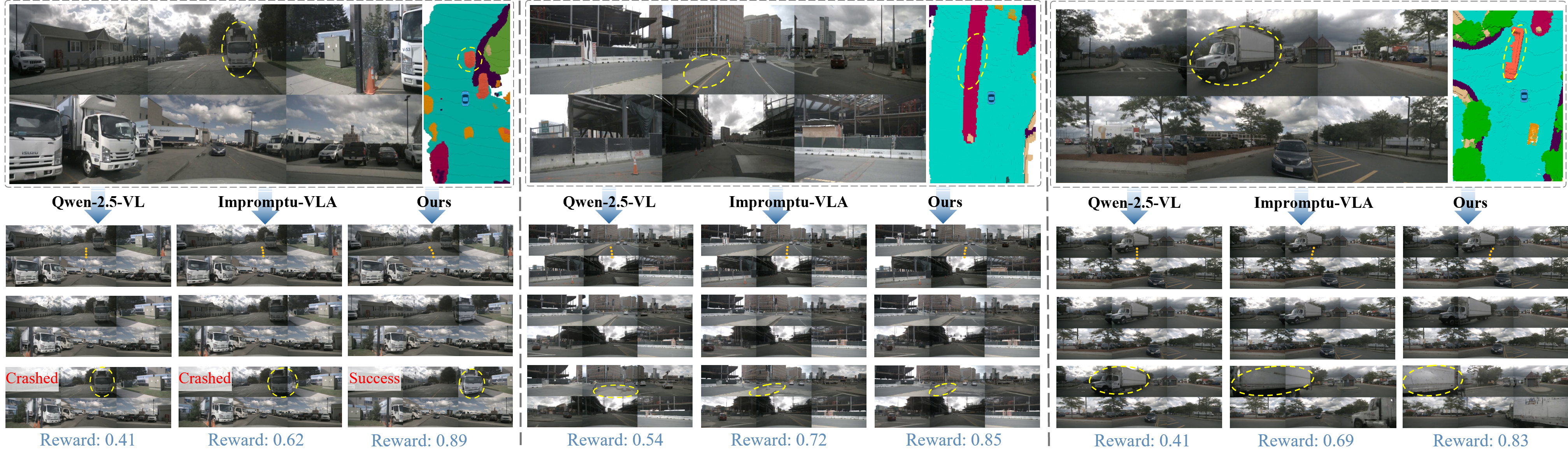}
     \vspace{-20pt}
    \caption{ 
    \textbf{Average rewards of different trajectories,} computed using the proposed 3D occupancy-grounded reward function. The rewards effectively evaluate the feasibility of planning trajectories in the presence of obstacles (\eg, the oncoming truck, drivable area boundaries of median strips).
    }
    \label{fig_reward}
     \vspace{-20pt}
\end{figure}

\noindent\textbf{Closed-loop Simulation with Semantic-Geometric Agents.}\label{sec_trajectory}
As illustrated in Fig.~\ref{fig_teaser1} (c) and Fig.~\ref{fig_overall}, OmniNWM functions as a differentiable environment simulator $\mathcal{E}$ that enables the closed-loop evaluation of planning agents. Formally, we establish a feedback loop where the agent policy $\pi$ observes the generated state $\mathcal{S}_t$ and outputs an action $\mathbf{a}_t$, which is then projected into our canonical Plücker manifold to condition the generation of next state $\mathcal{S}_{t+1} \sim \mathcal{E}(\mathcal{S}_t, \mathbf{a}_t)$.

\noindent\textit{OmniNWM-VLA Agent.} To validate this closed-loop pipeline and fully leverage the semantic and geometric richness of the PDiT generation outputs, we develop a specialized Vision-Language-Action (VLA) agent based on Qwen-VL~\cite{QwenVL}. Unlike previous planners~\cite{zheng2025diffusion,lee2023refining,yang2024diffusion} that rely solely on sparse objects or layouts, OmniNWM-VLA digests high-dimensional panoramic contexts. As shown in Fig.~\ref{fig_overall}, we introduce a plug-and-play Tri-Modal Mamba-based Interpreter (Tri-MMI) acting as a state projector $\phi: \{\texttt{RGB}, \texttt{Depth}, \texttt{Sem}\} \to \mathbb{R}^d$. This module efficiently integrates photometric, geometric, and semantic contexts via selective state-space modeling before tokenization, enabling {semantic-geometric reasoning} (\eg, ``yield to the truck on the left" in Fig.~\ref{fig_reward}).


\noindent\textit{High-Fidelity Action Space.} To match our canonical geometric control (Sec.~\ref{sec_raymap}), the agent's head outputs a dense trajectory tuple $(x, y, \theta)$ that includes heading angles. Unlike prior 2Hz methods~\cite{chi2025impromptu,zeng2025futuresightdrive}, our 12Hz closed-loop pipeline minimizes the sim-to-real gap, simulating reactive maneuvers (\eg, cut-in in Fig.~\ref{re_control}). A static trajectory bootstraps the initial state before the agent takes full control.\looseness=-1

\noindent\textit{Occupancy-Grounded Evaluation.} OmniNWM supports the integration of diverse agents as a robust evaluation environment. We evaluate OmniNWM-VLA against other baselines~\cite{chi2025impromptu,QwenVL} (all finetuned and adapted to 12Hz), using our intrinsic occupancy-grounded rewards in Fig.~\ref{fig_reward}. As illustrated in the figure, our reward function effectively discriminates policy quality in critical scenarios, such as navigating near median strips (column 2) or oncoming vehicles (columns 1 \& 3).\looseness=-1


\begin{figure}[!t]
\vspace{-3pt}
    \centering
    \includegraphics[width=0.95\linewidth]{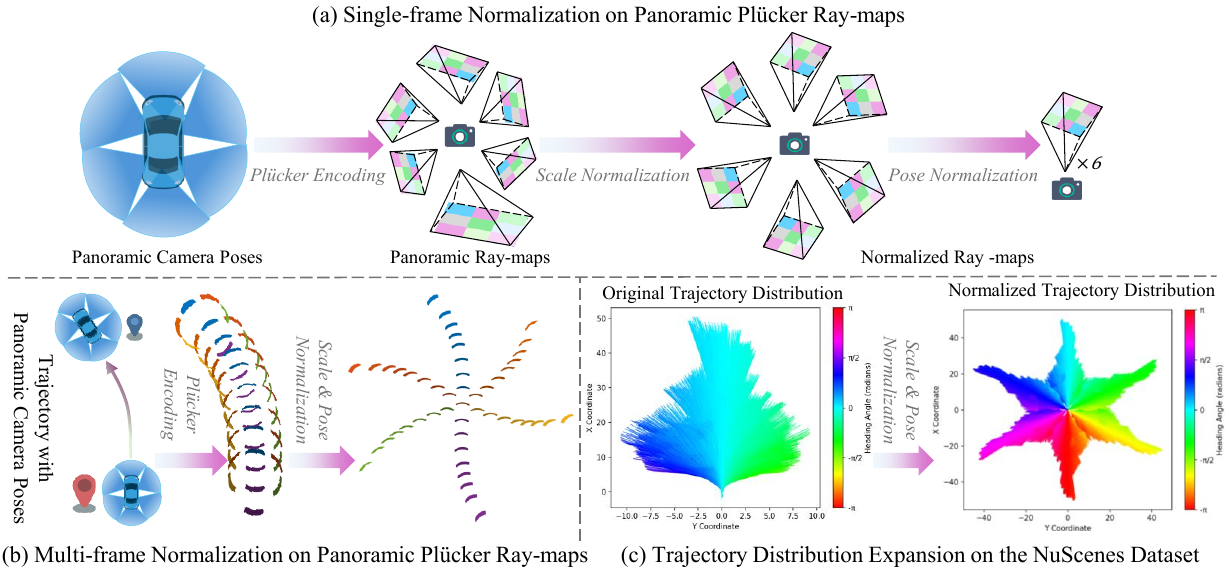}
    \vspace{-7pt}
    \caption{\textbf{Panoramic normalized ray-map encoding.}
    (a) 
    Plücker ray-maps derived from panoramic camera poses undergo pose and scale normalization. 
    (b) The normalization process constructs different trajectories within a unified, invariant 3D Plücker space. (c) Compared to the original nuScenes dataset, our strategy significantly enhances the diversity of trajectory distributions.
    }
    \label{fig_ray_normal}
    \vspace{-15pt}
\end{figure}

\vspace{-12pt}
\subsection{Canonical Encoding via Normalized Panoramic Ray-maps}\label{sec_raymap}
\vspace{-3pt}
Existing methods suffer from {geometric covariate shift} as current sparse representations (\eg, waypoints, camera poses)~\cite{he2024cameractrl,gao2024magicdrive,wang2023drivedreamer} entangle {ego-motion dynamics} with specific rig geometry, causing models to overfit to extrinsic calibrations and fail to generalize across varying sensor setups. 
To resolve this, our {Normalized Panoramic Plücker Ray-map} projects diverse sensor configurations into a unified {canonical Plücker manifold}. By decoupling scene dynamics from sensor geometry, we enable precise, zero-shot panoramic control across different camera rigs.

\noindent\textbf{Parameter-free Panoramic Plücker Encoding.}
While previous works are tailored for monocular sequences without multi-camera constraints~\cite{he2024cameractrl,he2025cameractrl,wang2024motionctrl}, our approach explicitly preserves multi-view consistency across camera rigs. As illustrated in Fig.~\ref{fig_overall}, we employ a parameter-free encoder to map input trajectories into a high-dimensional ray-maps. These ray-maps are downsampled and patchified into Plücker embedding tokens, which are then concatenated with diffusion latent tokens for 3D full-attention processing. This parameter-free design facilitates the injection of control signals as precise, pixel-aligned geometric constraints, rather than relying on learned semantic priors.
Formally, we construct the ray-map from camera rigs, and define the raw Plücker embedding:\looseness=-1 
\vspace{-6pt}
\begin{equation}
{p}_{u, v}=\left({o} \times \hat{{d}}_{u, v}, \hat{{d}}_{u, v}\right) \in \mathbb{R}^6,
\end{equation}
where ${o}$ is the camera center, and $\hat{{d}}_{u, v}$ is the direction vector from ${o}$ to the pixel.

\noindent\textbf{Canonical Projection for Geometric Invariance.}
Raw Plücker embeddings remain sensitive to the absolute scale and pose of specific data collection rigs. Thus, we decouple motion dynamics from rig-specific scale and pose via projecting all camera rays into a unified reference frame (\eg, initial front view).
As shown in Fig.~\ref{fig_ray_normal} (a), we first unproject pixels via source intrinsics $K_k$ to obtain explicitly unit-normalized world direction vectors:\looseness=-1

\vspace{-3pt}
\begin{equation}
\hat{d}^{(k)}_{u,v} = \frac{R_k K_k^{-1} [u, v, 1]^T}{\| R_k K_k^{-1} [u, v, 1]^T \|_2}. 
\end{equation}

Next, we perform a rigid transformation to map the geometry into the canonical reference coordinate. The camera center relative to the reference view is:
\vspace{-3pt}
\begin{equation}
o^{(k \to 0)} = R_0 R_k^{-1} (t_k - t_0), 
\end{equation}
where $t_k$ represents the optical center of camera $k$ in world coordinates.
The direction vector is similarly rotated into the reference orientation:
\vspace{-3pt}
\begin{equation}
\hat{d}^{(k \to 0)}_{u,v} = R_0 R_k^{-1} \hat{d}^{(k)}_{u,v}.
\end{equation}

The final {Normalized Panoramic Ray-map} is derived by re-computing the Plücker coordinates anchored to this unified canonical frame:
\vspace{-3pt}
\begin{equation}
\hat{p}^{(k)}_{u,v} = \left( {o}^{(k \to 0)} \times \hat{d}^{(k \to 0)}_{u,v}, \hat{d}^{(k \to 0)}_{u,v} \right), 
\end{equation}
which yields a geometrically consistent representation invariant to the recording rigs. Moreover, this invariant representation unifies multi-view trajectories into a shared 3D Plücker space (Fig.~\ref{fig_ray_normal}~(b)), significantly enriching trajectory distribution diversity to facilitate generalizable learning (Fig.~\ref{fig_ray_normal}~(c)).\looseness=-1

\begin{figure}[!t]
 \vspace{-1pt}
  \begin{scriptsize}
    \centering
    \includegraphics[width=0.95\linewidth]{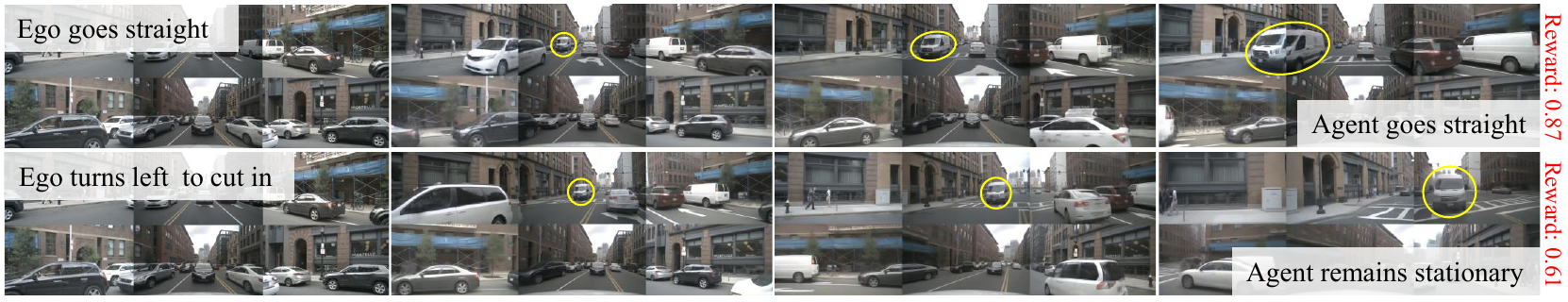}
     \vspace{-11pt}
    \caption{\textbf{Interactive Simulation.} Surrounding agents {reactively} yield to ego's cut-in.
    }
    \label{re_control}
    \vspace{-20pt}
     \end{scriptsize}
\end{figure}

\noindent\textbf{Emergent Interactive Dynamics via Equilibrium Modeling.}\label{sec_nash}
We distinguish between arbitrary {scenario editing} (``God-mode'' control) and {reactive world modeling} (simulating realistic environmental responses). While arbitrary control often violates physical consistency~\cite{li2025uniscene,gao2023magicdrive}, we argue that a valid world model should approximate {reactive environmental dynamics} by preserving the latent {Nash Equilibrium}~\cite{holt2004nash,daskalakis2009complexity} inherent in expert driving demonstrations.\looseness=-1

\noindent\textit{Equilibrium Manifold Hypothesis.} We posit that naturalistic driving datasets are not random collections of independent trajectories, but rather represent a multi-agent system in a momentary Nash Equilibrium. Let $\mathcal{A} = \{a_\texttt{ego}, a_{1}, \dots, a_{N}\}$ be the set of actions for ego-vehicle and $N$ surrounding agents. The joint action profile $\mathcal{A}^*$ satisfies the condition that no agent $i$ can unilaterally deviate without increasing their cost function $J_i$ (\eg, collision risk).
Thus, the training distribution support lies on this {Equilibrium Manifold}.\looseness=-1

\noindent\textit{World Modeling as Causal Reaction.} 
Consequently, forcing arbitrary trajectories for other agents ($\texttt{do}(a_{k}')$) pushes the state off this manifold, leading to physical inconsistencies.
OmniNWM is explicitly designed to model the conditional probability $P(\mathcal{S}_{t+1}|\mathcal{S}_{t}, a_\texttt{ego})$. By restricting explicit control to the ego-vehicle ($a_\texttt{ego}$) and treating surrounding agents as reactive latent variables, we ensure the model samples from the learned equilibrium distribution.
This design facilitates {interactive emergence}: as shown in Fig.~\ref{re_control}, when the ego-vehicle cuts in, the model generates a ``yielding'' behavior for the truck. This occurs not because of heuristic scripting, but because ``yielding'' is the highest-likelihood response on the equilibrium manifold conditioned on the ego's aggressive action.

\noindent\textbf{Mitigate Geometric Covariate Shift.}
Existing driving world models~\cite{gao2024magicdrive,gao2024magicdrivedit,guo2025dist} and monocular camera control methods~\cite{he2024cameractrl,he2025cameractrl,wang2024motionctrl} condition on raw extrinsics $\xi$, entangling intrinsic motion dynamics with sensor-specific disturbance factors. This creates a geometric covariate shift where the support of training and out-of-distribution (OOD) testing distributions can become disjoint~\cite{huang2025selfforcing,chen2025diffusion}, causing the KL divergence to be arbitrarily large: $D_{KL}(P_\texttt{tgt} || P_\texttt{src}) \gg 0$. 
This formulation elucidates the difficulty that previous methods face in zero-shot transfer.
Our normalized ray-map acts as a canonical projection operator $\Phi: \mathbb{R}^6 \to \mathcal{M}_\texttt{motion}$, filtering out rig-specific geometry to map diverse sensor configurations onto a shared canonical motion manifold. By aligning the manifold support (as visualized in Fig.~\ref{fig_ray_normal} (c)), we aim to mitigate the divergence:
\vspace{-6pt}
\begin{equation}
\min_{\Phi} D_\texttt{KL}\left( P_\texttt{tgt}(\Phi(\xi)) || P_\texttt{src}(\Phi(\xi)) \right) \approx \epsilon,
\end{equation}
where $\epsilon$ represents irreducible aleatoric uncertainty. 
This projection facilitates robust zero-shot generalization across distinct camera rigs, including novel trajectories(\eg, reversing in Fig.~\ref{fig_teaser1} (b)), unseen datasets (\eg, 200 frames on nuPlan in Fig.~\ref{fig_abl_forcing} (b)) and novel camera rigs (\eg, 3-view and 6-view in Fig.~\ref{fig_zero_shot}), capabilities that remain challenging for previous monocular adaptations~\cite{he2024cameractrl,he2025cameractrl,wang2024motionctrl}.\looseness=-1

\vspace{-10pt}
\subsection{Covariate Shift Mitigation via Structured Panoramic Forcing}\label{sec_forcing}
\vspace{-5pt}
Previous methods typically suffer from {covariate shift}~\cite{li2025occscene,gao2024magicdrive,guo2025dist}: conditioning on GT history during training but error-prone predictions during inference causes distribution drift, which accumulates errors and ultimately collapses generated sequences. Therefore, our structured panoramic forcing systematically corrupts training contexts with structured noise. This stochastic regularization aligns training and inference distributions by forcing the model to learn a {restorative mapping}, effectively projecting drifting states back onto the valid data manifold.\looseness=-1

\begin{figure}[!t]
\vspace{-3pt}
\centering
\includegraphics[width=0.99\linewidth]{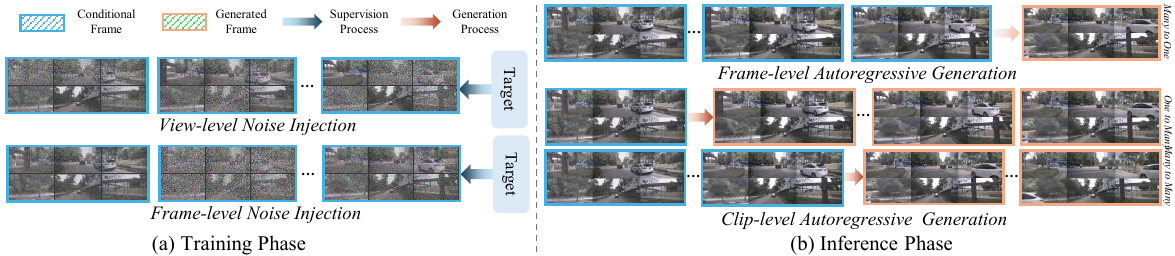}
\vspace{-10pt}
\caption{\textbf{Structured panoramic forcing strategy.}
(a) During training, independent multi-level noise is injected along view-wise and frame-wise dimensions.
(b) During inference, flexible and robust generation is enabled via either frame-level autoregression (many-to-one) or clip-level autoregression (one-to-many or many-to-many). }
\label{fig_forcing}
\vspace{-20pt}
\end{figure}

\noindent\textbf{Structured Noise Injection.}
Unlike standard Gaussian noise that assumes independent and identically distributed errors, we model the error distribution in panoramic video generation as structurally coupled. Errors tend to be correlated either temporally (accumulated motion drift) or spatially (inter-view inconsistency).
Therefore, we introduce a decoupled noise injection scheme during training (Fig.~\ref{fig_forcing} (a)). For a latent representation $\mathbf{z}^{(t,v)}$ at frame $t$ and view $v$, we construct a corrupted context $\tilde{\mathbf{z}}^{(t,v)}$ by injecting independent hierarchical noise:\looseness=-1

\vspace{-6pt}
\begin{equation}
\tilde{\mathbf{z}}^{(t,v)} = \mathbf{z}^{(t,v)} + \underset{{\text{Temporal Drift}}}{\underbrace{\alpha^{(t)} \cdot \mathbf{\epsilon}_{\texttt{temp}}}} + \underset{{\text{Spatial Inconsistency}}}{\underbrace{\beta^{(v)} \cdot \mathbf{\epsilon}_{\texttt{spat}}}},
\end{equation}

\noindent where $\mathbf{\epsilon}_{\texttt{temp}} , \mathbf{\epsilon}_{\texttt{spat}} \sim \mathcal{N}(0, I)$ are independent noise vectors, and $\alpha^{(t)}, \beta^{(v)}$ are scaling factors. This structured formulation explicitly simulates the two dominant failure modes in long-horizon generation: $\mathbf{\epsilon}_{\texttt{temp}}$ mimics the trajectory drift over time, while $\mathbf{\epsilon}_{\texttt{spat}}$ mimics the geometric misalignment between camera views.

\noindent\textbf{Flexible Inference Modes.}
This robust training paradigm naturally enables flexible inference capabilities, as the model becomes resilient to varying qualities of conditioning context.
As shown in Fig.~\ref{fig_forcing} (b), we support:
(1) {Frame-level Autoregression}: $\mathbf{\hat{x}}_{t+1} = f_{\theta}(\tilde{\mathbf{x}}_{t-K:t})$, optimal for high-precision trajectory planning.
(2) {Clip-level Autoregression}: $\mathbf{\hat{x}}_{t+1:t+L} = f_{\theta}(\tilde{\mathbf{x}}_{t}/\tilde{\mathbf{x}}_{t-M:t})$, which trades temporal granularity for computational efficiency in long-term generations.
Evaluation results with clip-level autoregression in Fig.~\ref{fig_abl_forcing} (b) (200-frame comparison) and Tab.~\ref{tab_abl_forcing} (386.72 to 25.22 FVD) demonstrate the effectiveness of our strategy as a structural prerequisite for stable world modeling.\looseness=-1

\noindent\textbf{Alleviate Exposure Bias via Manifold Thickening.}
Previous methods learns a transition operator $\mathcal{T}$ valid only on the thin GT manifold $\mathcal{M}_\texttt{GT}$, and thereby suffer from {exposure bias}~\cite{li2025occscene,gao2024magicdrive,guo2025dist}. During inference, inevitable approximation errors cause the trajectory to drift off-manifold ($\mathbf{x}_t \notin \mathcal{M}_\texttt{GT}$), entering undefined regions where errors compound exponentially.
Structured Panoramic forcing addresses this by {Stochastic Manifold Thickening}. By training on perturbed states, we expand the valid support to a tubular neighborhood $\mathcal{M}_\epsilon = \{\mathbf{x}:d(\mathbf{x}, \mathcal{M}_\texttt{GT}) < \epsilon\}$. Crucially, this encourages the learned transition operator $\mathcal{T}$ to approximate a local {contractive property}:
$
|\mathcal{T}(\tilde{\mathbf{x}}) - \mathcal{T}(\mathbf{x})| \leq \lambda |\tilde{\mathbf{x}} - \mathbf{x}|, \  \text{where } \lambda < 1.
$
This formulation aims to stabilize the generative process, treating the true data manifold $\mathcal{M}_\texttt{GT}$ as an {attractor}. Perturbations are dampened rather than amplified, which facilitates bounded error accumulation for long-horizon rollouts.\looseness=-1

\vspace{-10pt}
\section{Experiments}
\vspace{-8pt}

\subsection{Experimental Setup}
\vspace{-6pt}
Our model is trained on nuScenes~\cite{caesar2020nuscenes} and nuScenes-Occupancy~\cite{wang2023openoccupancy} datasets. 
Training is conducted on 48 NVIDIA A800 GPUs with a batch size of 48. We employ the AdamW optimizer~\cite{loshchilov2017decoupled} with a learning rate of $1\times10^{-4}$ and weight decay of 0.01.
To ensure stable convergence, we adopt a progressive three-stage training protocol: (1) Single-view Control. The model is first trained on single-view sequences (17 frames, $224\times400$ resolution) for 10k iterations. (2) Multi-view Joint Generation. We extend training to joint 6-view panoramic generation for 3k iterations. (3) Variable-length Fine-tuning. We introduce variable sequence lengths (17 or 33 frames) and resolutions (up to $448\times800$) for a final 3k iterations to enhance adaptability.
Additional details are provided in the supplementary.\looseness=-1

\begin{table}[!t]
\vspace{-3pt}
\begin{center}
\caption{\textbf{Quantitative evaluation} of RGB video generation on nuScenes validation set.
Our method outperforms SOTAs without heavy volumetric input conditions.
}
 \vspace{-10pt}
\scriptsize
\vspace{-0pt}
\renewcommand\tabcolsep{10.0pt}
\centering
\resizebox{0.7\linewidth}{!}{
\begin{tabular}{l|ccc|cc}
\toprule Method & \makecell[c]{Volumetric\\ Condition-Free} & Multi-view & Video & FID $\downarrow$ & FVD $\downarrow$ \\ \midrule
DriveGAN~\cite{kim2021drivegan} & \textcolor{ForestGreen}{\usym{2713}} & \textcolor{red}{\usym{2717}} & \textcolor{ForestGreen}{\usym{2713}} & 73.40 & 502.30 \\
BEVControl~\cite{yang2023bevcontrol} & \textcolor{ForestGreen}{\usym{2713}} & \textcolor{ForestGreen}{\usym{2713}} & \textcolor{red}{\usym{2717}} & 24.85 & - \\
DriveDreamer~\cite{wang2023drivedreamer} & \textcolor{ForestGreen}{\usym{2713}} & \textcolor{red}{\usym{2717}} & \textcolor{ForestGreen}{\usym{2713}} & 14.90 & 340.80 \\
BEVGen~\cite{swerdlow2024streetview} & \textcolor{ForestGreen}{\usym{2713}} & \textcolor{ForestGreen}{\usym{2713}} & \textcolor{red}{\usym{2717}} & 25.54 & - \\
DrivingGPT~\cite{chen2024drivinggpt} & \textcolor{ForestGreen}{\usym{2713}} & \textcolor{red}{\usym{2717}} & \textcolor{ForestGreen}{\usym{2713}} & 12.78 & 142.61  \\
Vista~\cite{gao2025vista} & \textcolor{ForestGreen}{\usym{2713}} & \textcolor{red}{\usym{2717}} & \textcolor{ForestGreen}{\usym{2713}}& 6.90 & 89.40 \\ 
DrivingWorld~\cite{hu2024drivingworld} & \textcolor{ForestGreen}{\usym{2713}} & \textcolor{red}{\usym{2717}} & \textcolor{ForestGreen}{\usym{2713}} & 7.40 & 90.90 \\
Epona~\cite{zhang2025epona} & \textcolor{ForestGreen}{\usym{2713}} & \textcolor{red}{\usym{2717}} & \textcolor{ForestGreen}{\usym{2713}} & 7.50 & 82.80 \\
\midrule
WoVoGen~\cite{lu2023wovogen} & \textcolor{red}{\usym{2717}} &  \textcolor{ForestGreen}{\usym{2713}} & \textcolor{ForestGreen}{\usym{2713}} & 27.60 & 417.70 \\
X-Scene~\cite{yang2025x} & \textcolor{red}{\usym{2717}} &  \textcolor{ForestGreen}{\usym{2713}} &  \textcolor{red}{\usym{2717}} & 11.29 & -  \\
OccScene~\cite{li2025occscene}& \textcolor{red}{\usym{2717}} & \textcolor{ForestGreen}{\usym{2713}} &\textcolor{ForestGreen}{\usym{2713}} & 11.87 & -\\
UniScene~\cite{li2025uniscene}& \textcolor{red}{\usym{2717}}  & \textcolor{ForestGreen}{\usym{2713}}&  \textcolor{ForestGreen}{\usym{2713}} & {6.45} & {71.94} \\ 
DiST-4D~\cite{guo2025dist}& \textcolor{red}{\usym{2717}} & \textcolor{ForestGreen}{\usym{2713}}&  \textcolor{ForestGreen}{\usym{2713}}& {7.40} & {25.55} \\
\midrule

MagicDrive~\cite{gao2023magicdrive} & \textcolor{ForestGreen}{\usym{2713}} & \textcolor{ForestGreen}{\usym{2713}} & \textcolor{ForestGreen}{\usym{2713}} & 16.20 & - \\  

Panacea~\cite{wen2024panacea} & \textcolor{ForestGreen}{\usym{2713}} & \textcolor{ForestGreen}{\usym{2713}} & \textcolor{ForestGreen}{\usym{2713}} & 16.96 & 139.00\\
Drive-WM~\cite{wang2023drivewm} & \textcolor{ForestGreen}{\usym{2713}}& \textcolor{ForestGreen}{\usym{2713}} & \textcolor{ForestGreen}{\usym{2713}} &  15.80  &  122.70  \\
GenAD~\cite{zheng2024genad} & \textcolor{ForestGreen}{\usym{2713}} & \textcolor{ForestGreen}{\usym{2713}} & \textcolor{ForestGreen}{\usym{2713}} & 15.40 & 184.00 \\ 
DriveDreamer-2~\cite{zhao2024drivedreamer2}& \textcolor{ForestGreen}{\usym{2713}}  & \textcolor{ForestGreen}{\usym{2713}}&  \textcolor{ForestGreen}{\usym{2713}} & 11.20 & {55.70} \\
MagicDrive-V2~\cite{gao2024magicdrive} & \textcolor{ForestGreen}{\usym{2713}} & \textcolor{ForestGreen}{\usym{2713}}&  \textcolor{ForestGreen}{\usym{2713}} & 20.91 & 94.84 \\
\midrule
\rowcolor{gray!10} OmniNWM & \textcolor{ForestGreen}{\usym{2713}}
& \textcolor{ForestGreen}{\usym{2713}}&  \textcolor{ForestGreen}{\usym{2713}}& \textbf{5.45}  & \textbf{23.63}  \\ 
\bottomrule
\end{tabular}
 }
\vspace{-17pt}
\label{tab_video}
\end{center}
\end{table}

\begin{table}[!t]
\centering
\caption{
\textbf{Quantitative evaluation} of camera control accuracy on nuScenes validation set. 
MotionCtrl* and CameraCtrl* are adapted and fine-tuned for fairness.
Our method outperforms SOTAs with panoramic Plücker ray-map normalization (`Norm'). 
} 
\vspace{-10pt}
\renewcommand\tabcolsep{5.0pt}
\resizebox{0.9\columnwidth}{!}{
\begin{tabular}{l|cccccccc}
\toprule
\multirow{2}{*}{Method}  & \multirow{2}{*}{Norm}   & \multirow{2}{*}{RotErr (radian) $\downarrow$} & \multirow{2}{*}{TransErr (meter) $\downarrow$}   & \multicolumn{2}{c}{VelErr (m/s) $\downarrow$}   & \multicolumn{2}{c}{AccErr (m/s$^{2}$ ) $\downarrow$}  \\ \cmidrule(lr){5-8}
 && & & x-axis  & y-axis & x-axis  & y-axis    \\  
\midrule
MotionCtrl*~\cite{wang2024motionctrl}  & - & 2.36  & 12.71 & 2.55 & 0.28 & 50.47 & 3.68 \\
CameraCtrl*~\cite{he2024cameractrl} & -  & 1.67  & 9.48 &  1.89 & 0.16 & 35.72 & 2.94  \\
PosePilot~\cite{jin2025posepilot}+Vista~\cite{gao2025vista}   & - & 1.53  & 6.52 &  1.65 & 0.14 & 26.86 & 1.97     \\
UniScene~\cite{li2025uniscene}  & - & 1.62 & 7.56 & 1.77 & 0.15 & 30.76 & 2.27 \\ \midrule
OmniNWM  & \textcolor{red}{\usym{2717}}  & 1.71 & 9.75 & 1.91  & 0.17 & 33.12 &  2.46  \\
\rowcolor{gray!10} OmniNWM & \textcolor{ForestGreen}{\usym{2713}} & \textbf{0.16} & \textbf{1.18}  & \textbf{0.72}  & \textbf{0.07} & 11.84 & \textbf{1.06}  
\\
\bottomrule
\end{tabular}
}
\label{table_camctrl}
\vspace{-18pt}
\end{table}

\vspace{-6pt}
\subsection{Main Results}\label{main_results}
\vspace{-5pt}
\noindent\textbf{Video Generation Quality.}
 We evaluate generation quality using FID~\cite{heusel2017gans} and FVD~\cite{unterthiner2018towards} at $224\times400$ resolution with 17-frame clips, following~\cite{gao2024magicdrive,guo2025dist}. 
As shown in Tab.~\ref{tab_video}, our method achieves SOTA performance (5.45 FID, 23.63 FVD) without relying on heavy {volumetric input conditions} (\eg, voxel grids~\cite{lu2023wovogen,yang2025x,li2025occscene,li2025uniscene} or point clouds~\cite{guo2025dist}), conditioned on efficient normalized panoramic ray-maps.
We further evaluate the generated panoramic depth maps using standard metrics~\cite{long2021multi,zou2024m2depth} against LiDAR-projected GT~\cite{guo2025dist}. 
As shown in Tab.~\ref{tab_depth_results}, our method obviously outperforms the generative method of Dist-4D~\cite{guo2025dist} and surpasses discriminative baselines~\cite{wei2023surrounddepth,zou2024m2depth} limited by generalization. \looseness=-1

\begin{figure}[!t]
    \centering
    \vspace{-1pt}
\includegraphics[width=0.7\linewidth]{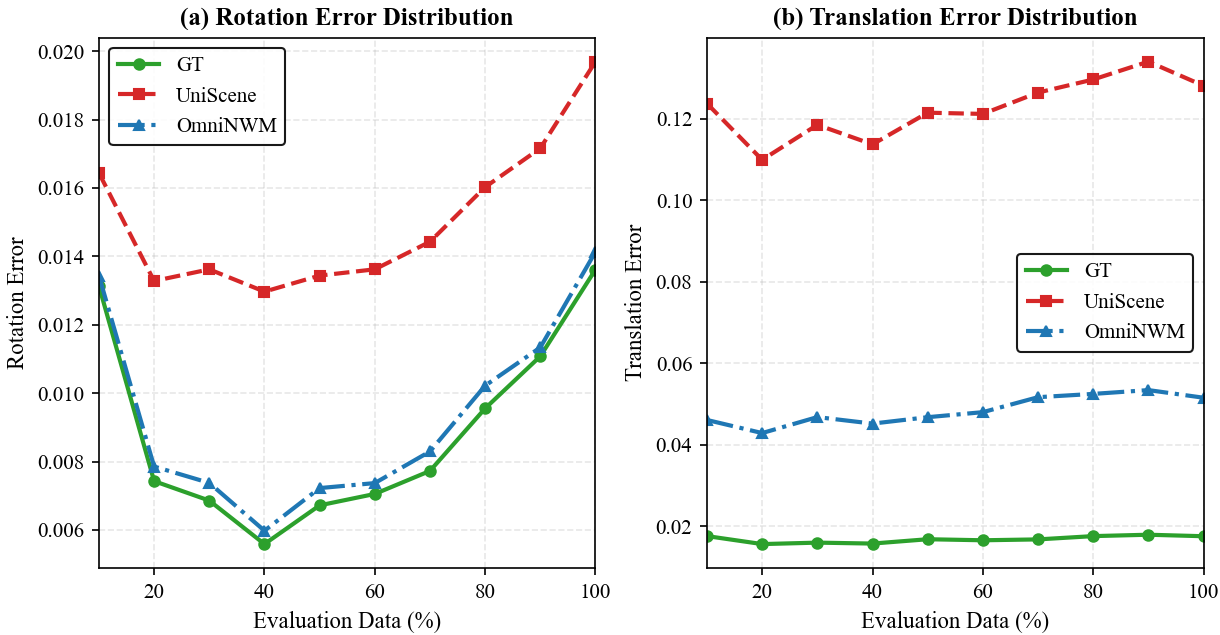}
    \vspace{-8pt}
    \caption{
     \textbf{Distribution} of camera control performance on nuScenes validation set.
    }
    \label{fig_camctrl}
    \vspace{-10pt}
\end{figure}

\noindent\textbf{Camera Control Accuracy.} 
We implement camera control evaluation following previous works~\cite{he2024cameractrl,jin2025posepilot}: camera poses are recovered from generated videos via COLMAP~\cite{schonberger2016structure} and aligned to GT trajectories using Sim(3) alignment. We measure Rotation/Translation Error (RotErr/TransErr) and dynamic consistency (VelErr/AccErr) across the full 150-scene validation set for all methods.
As shown in Tab.~\ref{table_camctrl} and Fig.~\ref{fig_camctrl}, OmniNWM significantly reduces drift (1.18 m TransErr vs. 7.56 m for UniScene~\cite{li2025uniscene}). The superior dynamic error metrics confirm that our Normalized Plücker Ray-maps decouple motion from rig geometry, enabling precise control with the canonical geometric action encoding scheme.\looseness=-1

\begin{table}[!t]
\vspace{-0pt}
\caption{
\textbf{Quantitative evaluation} of semantic occupancy on nuScenes-Occupancy. 
}
\vspace{-11pt}
\renewcommand\tabcolsep{15.0pt}
	\centering
   \resizebox{0.7\linewidth}{!}{
	\begin{tabular}{l |c |c | c  c}
    \toprule
 Category & Method  & Input & \makecell[c]{IoU}  $\uparrow$ & \makecell[c]{mIoU}  $\uparrow$ \\
    \midrule
\multirow{10}{*}{Discriminative} & 
AICNet$^\text{*}$~\cite{li2020anisotropic}  & Camera\&Depth  & 23.8 &10.6   \\ 
&  3DSketch$^\text{*}$~\cite{chen20203d}  & Camera\&Depth & {25.6} &10.7  \\ 
    \cline{2-5}
& LMSCNet~\cite{roldao2020lmscnet} & LiDAR & {27.3} & 11.5  \\
& JS3C-Net~\cite{yan2021sparse} & LiDAR & 30.2 & 12.5 \\
& L-CONet~\cite{wang2023openoccupancy} & LiDAR & {30.9} & {15.8} \\
    \cline{2-5}
&	MonoScene~\cite{cao2022monoscene} & Camera  & 18.4 & 6.9  \\
&  	TPVFormer~\cite{huang2023tri} & Camera & 15.3 & 7.8  \\ 
&     C-CONet~\cite{wang2023openoccupancy} & Camera & 20.1 & 12.8 \\
&   HTCL~\cite{li2024hierarchical} & Camera & 21.4 & 14.1 \\

&     SparseOcc~\cite{Tang_2024_CVPR} & Camera  & 21.8 & 14.1 \\ 
&  Hi-SOP~\cite{li2026hierarchical} & Camera & 24.5 & 16.4 \\
\midrule
\multirow{3}{*}{Generative} 
& OccScene~\cite{li2025occscene} & Camera & - & 12.2 \\
&  OccGen~\cite{wang2024occgen} & Camera & 23.4 & {14.5}  \\
    \cline{2-5}

 & \cellcolor{gray!10} OmniNWM &  \cellcolor{gray!10} Camera &  \cellcolor{gray!10} \textbf{33.3} & \cellcolor{gray!10} \textbf{19.8}  \\  
\bottomrule
\end{tabular}}
\label{table_occ_nuscene}
\vspace{-12pt}
\end{table}

\begin{table}[!t]
\centering
\caption{\textbf{Quantitative evaluation} of panoramic depth on nuScenes validation set.}
\vspace{-12pt}
\renewcommand\tabcolsep{12.0pt}
\resizebox{0.7\linewidth}{!}{
\begin{tabular}{l|c|ccccc}
\toprule
 Category & {Method} &  Abs. Rel. $\downarrow$ & $\delta < 1.25$ $\uparrow$ & $\delta < 1.25^2$ $\uparrow$  \\
\midrule
\multirow{2}{*}{Discriminative} & SurroundDepth~\cite{wei2023surrounddepth} &   0.28  &0.66 &  0.84 \\
& M$^2$Depth~\cite{zou2024m2depth}        &  {0.26}  & {0.73} & {0.87}  \\ 
\midrule
\multirow{2}{*}{Generative}  & Dist-4D~\cite{guo2025dist}  &  0.39  &  0.58 &  0.81 \\
 & \cellcolor{gray!10} OmniNWM  & \cellcolor{gray!10}\textbf{0.23}   & \cellcolor{gray!10}\textbf{0.81} & \cellcolor{gray!10}\textbf{0.93}   \\   
\bottomrule
\end{tabular}}
\label{tab_depth_results}
\vspace{-12pt}
\end{table}

\noindent\textbf{Occupancy Prediction Quality.}
We evaluate occupancy prediction using IoU and mIoU following~\cite{li2025uniscene,li2025occscene}. As shown in Tab.~\ref{table_occ_nuscene}, OmniNWM achieves a SOTA 19.8 mIoU, outperforming Camera-Depth baselines~\cite{li2020anisotropic,chen20203d} and LiDAR-based methods~\cite{roldao2020lmscnet,yan2021sparse,wang2023openoccupancy}. 
Despite lifting from generated 2D observations rather than raw sensor data, our method recovers 3D structures comparable to LiDAR-based perception~\cite{roldao2020lmscnet,yan2021sparse,wang2023openoccupancy}. 
To ensure fairness, our occupancy generator is supervised on dense annotations~\cite{wang2023openoccupancy} following previous camera-based discriminative baselines~\cite{li2024hierarchical,li2026hierarchical}.
These results demonstrate that our geometric lifting strategy effectively preserves 3D structural consistency with the joint generation scheme.\looseness=-1

\begin{figure}[!t]
\vspace{-0pt}
\centering
\includegraphics[width=0.7\linewidth]{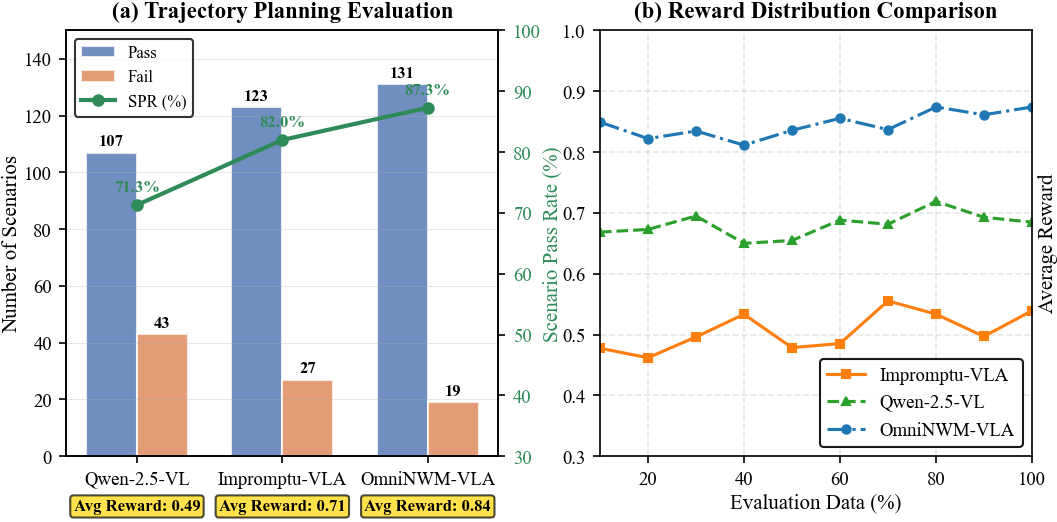}
\vspace{-8pt}
\caption{
\textbf{(a) Closed-loop evaluation} of different planners. \textbf{(b) Distribution} of rewards across different planners on nuScenes validation set.
}
\label{fig_vla_spr_compare}
\vspace{-11pt}
\end{figure}

\noindent\textbf{Trajectory Planning Evaluation.}
Fig.~\ref{fig_vla_spr_compare}~(a) presents closed-loop evaluations for trajectory planning on pass/fail counts, and Scenario Pass Rate (SPR)~\cite{sun2021scenario} across 150 nuScenes validation scenes. To rigorously isolate planner capability from environmental factors, all baselines~\cite{chi2025impromptu,QwenVL} and our OmniNWM-VLA are evaluated under identical textual prompts at 12 Hz. 
As shown in the figure, OmniNWM-VLA surpasses Impromptu-VLA~\cite{chi2025impromptu} and Qwen-2.5-VL~\cite{QwenVL} with 87.3\% SPR. 
Furthermore, Fig.~\ref{fig_vla_spr_compare}~(b) illustrates the trajectory reward distributions using our occupancy-grounded metric. 
The distinct separation of these distributions confirms that our reward function reliably differentiates policy performance, validating its utility for closed-loop evaluation.

\begin{figure}[!t]
    \centering
        \vspace{-0pt}
\includegraphics[width=0.9\linewidth]{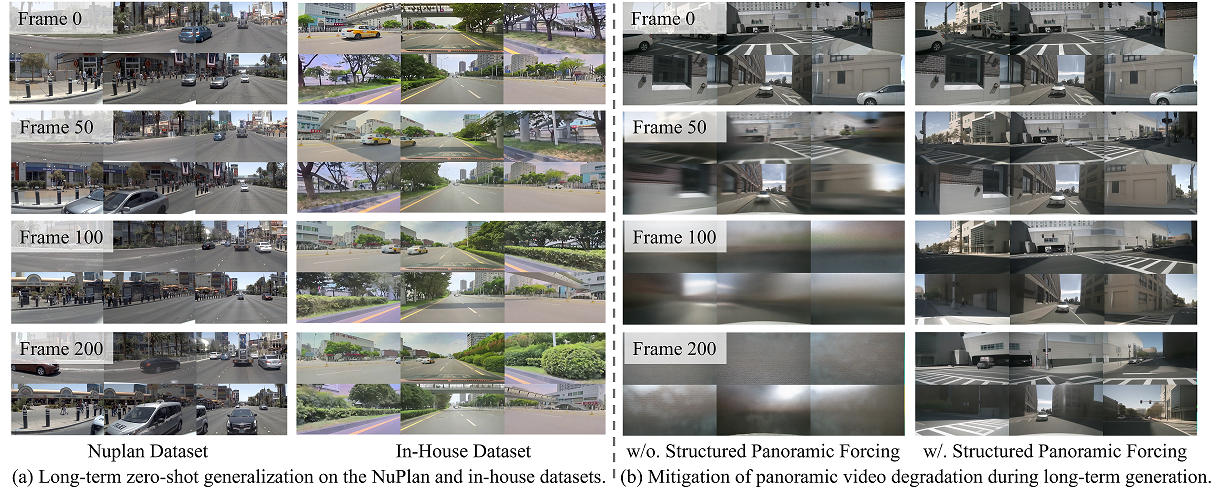}
    \vspace{-10pt}
    \caption{
    \textbf{Qualitative evaluation} of long-term generation. (a) Long-term zero-shot generalization on nuPlan and in-house datasets. (b) Effective mitigation of degradation during long-term generation with our structured panoramic forcing. 
    }
    \label{fig_abl_forcing}
    \vspace{-16pt}
\end{figure}

\begin{table}[!t]
\vspace{-0pt}
\caption{\textbf{Ablation study} on the structured panoramic forcing strategy. }
\vspace{-12pt}
\centering
\renewcommand\tabcolsep{15.0pt} 
\resizebox{0.9\columnwidth}{!}{ 
\begin{tabular}{l|ccccc}
\toprule
Method & FVD$_{17}$ & FVD$_{33}$ & FVD$_{65}$ & FVD$_{129}$ & FVD$_{201}$ \\ \midrule
No Structured Panoramic Forcing (\ie, pure autoregression) & 26.79 & 59.64 & 102.79 & 249.74 & 386.72 \\
Standard Scheduled Sampling (\ie, unstructured noise) & 25.10 & 36.82 & 58.45 & 115.30 & 178.65 \\
\rowcolor{gray!10} {Structured Panoramic Forcing (ours)} & \textbf{23.63} & \textbf{24.14} & \textbf{24.72} & \textbf{25.05} & \textbf{25.22} \\
\bottomrule
\end{tabular}
}
\label{tab_abl_forcing}
\vspace{-23pt}
\end{table}

\begin{minipage}[!t]{0.39\linewidth}
    \centering  
    \vspace{-5pt}
    \scriptsize
    \vspace{-5pt}
    \captionof{table}{\textbf{Quantitative evaluation} of zero-shot generalization on the nuPlan validation set.}
    \renewcommand\tabcolsep{2pt}
    \resizebox{0.98\linewidth}{!}{ 
    \centering
\begin{tabular}{l|cccc}
\toprule
Method & FID$\downarrow$ & FVD$\downarrow$ & {RotErr$\downarrow$} &{TransErr$\downarrow$} \\ \midrule
Vista~\cite{gao2025vista} & 15.72  &  151.76 & 1.84 & 9.21 \\
MagicDrive~\cite{gao2023magicdrive}     &  26.74 &  295.47 & 2.65 & 13.50 \\ 
UniScene~\cite{li2025uniscene} & 15.54  & 147.69  & 1.92 & 8.84 \\
Epona~\cite{zhang2025epona}  &  17.62  &  168.72  & 2.01 & 9.45 \\
DiST-4D~\cite{guo2025dist} &  12.81 &  118.60  & 1.45 & 7.12 \\ 
\rowcolor{gray!10} OmniNWM  &  \textbf{9.51} &  \textbf{79.24}  & \textbf{0.28} & \textbf{1.65} \\
 \bottomrule
\end{tabular}
    }

    \label{re_quantitative_zero}
\end{minipage}
\hfill  
\begin{minipage}[!t]{0.57\linewidth}
\vspace{1pt}
    \centering  
    \scriptsize
\includegraphics[width=1.0\linewidth]{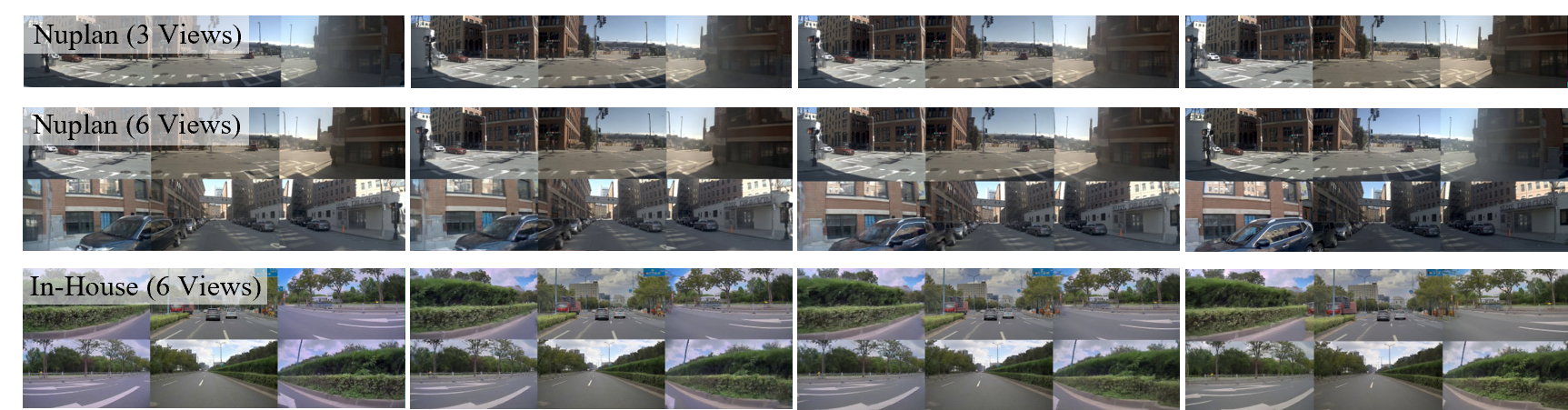}  
    \vspace{-12pt}
    \captionof{figure}{\textbf{Qualitative evaluation} of zero-shot generalization across different datasets and camera rig configurations.\looseness=-1}
    \label{fig_zero_shot}
\vspace{-0pt}
\end{minipage}

\begin{minipage}[t]{0.37\linewidth}
\vspace{2.0pt}
    \centering  
    \scriptsize
    \vspace{-5pt}
    \captionof{table}{\textbf{Ablation study} on the occupancy prediction inputs of generated RGB, semantic, and depth maps.}
    \renewcommand\tabcolsep{4pt}
    \resizebox{\linewidth}{!}{ 
    \centering
\begin{tabular}{ccc|cc}
\toprule
RGB & Semantics & Depth  & IoU $\uparrow$  & mIoU $\uparrow$   \\ \midrule
\textcolor{ForestGreen}{\usym{2713}} & \textcolor{ForestGreen}{\usym{2713}} &  &28.9 & 17.1 \\  
\textcolor{ForestGreen}{\usym{2713}} &  & \textcolor{ForestGreen}{\usym{2713}} &31.5 & 16.8 \\ \midrule
\rowcolor{gray!10}  \textcolor{ForestGreen}{\usym{2713}} & \textcolor{ForestGreen}{\usym{2713}} & \textcolor{ForestGreen}{\usym{2713}} & 33.3 & 19.8 \\  
\bottomrule
\end{tabular}
    }
    \label{table_ab_occ}
\end{minipage}
\hfill  
\begin{minipage}[t]{0.57\linewidth}
\vspace{12pt}
    \centering  
    \scriptsize
    \vspace{-10.0pt}
\includegraphics[width=0.95\linewidth]{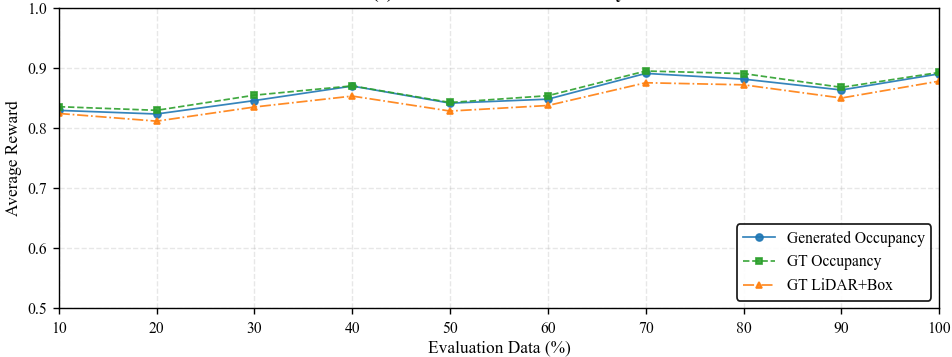}  
    \vspace{-10pt}
    \captionof{figure}{\textbf{Correlations} between the generated and GT occupancy on reward distributions.}
    \label{re_short_range_occ}
\end{minipage}

\noindent\textbf{Zero-shot Generalization.}
We evaluate OmniNWM without any fine-tuning across different datasets (\eg, nuPlan and in-house datasets) in Fig.~\ref{fig_abl_forcing} (a) and camera rigs (\eg, 3 and 6 views) in Fig.~\ref{fig_zero_shot}. 
Leveraging panoramic normalized ray-map encoding, our method demonstrates strong generalization across these diverse settings. 
Tab.~\ref{re_quantitative_zero} details the quantitative performance on 17-frame nuPlan clips following~\cite{li2025uniscene,li2025scaling}.
Crucially, we report {camera control accuracy} to assess cross-rig robustness.
While baselines suffer from geometric covariate shift and yield high trajectory errors (>7 m TransErr) due to overfitting nuScenes intrinsics, OmniNWM leverages normalized ray-maps to maintain precise control with {0.28 rad RotErr} and {1.65 m TransErr}, demonstrating that our canonical representation effectively decouples motion dynamics from sensor geometry.

\vspace{-6pt}
\subsection{Ablation Study}\label{sec_ablation}
\vspace{-3pt}
\noindent\textbf{Impact of Occupancy Prediction.} 
Tab.~\ref{table_ab_occ} ablates the occupancy module inputs. Joint semantic and depth generation yields significant gains of 3.0 and 2.7 mIoU, respectively.
To address circularity concerns in closed-loop evaluation, we quantify the alignment between rewards derived from our generated occupancy with GT occupancy or LiDAR+Box data with {average pearson correlations of ${r_\texttt{GT}=0.96}$ and ${r_\texttt{LiDAR}=0.94}$} (Fig.~\ref{re_short_range_occ}).
This alignment demonstrates the effectiveness of our occupancy generation approach with the geometric lifting strategy.\looseness=-1

\noindent\textbf{Impact of Canonical Panoramic Normalization.} 
Tab.~\ref{table_camctrl} evaluates the critical role of our panoramic ray-map normalization. Without the canonical normalization, the model suffers from geometric overfitting to specific sensor rigs ($1.71$ rad and $9.75$ m).
By projecting inputs onto the canonical Plücker manifold, we resolve scale and coordinate ambiguities, reducing rotation and translation errors by 90.6\% ($0.16$ rad) and 87.9\% ($1.18$ m).
Crucially, this invariance is necessary for zero-shot generalization, enabling stable 200-frame generation on the unseen nuPlan and in-house datasets (Fig.~\ref{fig_abl_forcing} (a) and Fig.~\ref{fig_zero_shot}).\looseness=-1

\noindent\textbf{Impact of Structured Panoramic Forcing.} 
Tab.~\ref{tab_abl_forcing} and Fig.~\ref{fig_abl_forcing} (b) evaluate our strategy against pure autoregression (without structured panoramic forcing) and standard scheduled sampling baselines~\cite{bengio2015scheduled,li2025uniscene}. While standard sampling (unstructured Gaussian noise) improves upon pure autoregression, it still suffers severe degradation over long horizons ($178.65$ FVD at $201$ frames) because it fails to model the structurally coupled spatial and temporal errors inherent in panoramic video (Fig.~\ref{fig_abl_forcing} (b)). 
 Moreover, our proposed structured panoramic forcing is essential to accurately thicken the training manifold, reducing compounding errors and maintaining a highly stable $25.22$ FVD over $201$ frames.\looseness=-1

\vspace{-10pt}
\section{Conclusion} 
\vspace{-6pt}
This paper presents OmniNWM, a unified framework that addresses the fragmentation in autonomous driving world models by approximating the joint multimodal posterior of the state-action-reward triad. By enforcing cross-modal consistency, we address the {modality drift} prevalent in modular architectures. Crucially, our canonical geometric action encoding decouples motion dynamics from sensor intrinsics with the normalized panoramic ray-maps, enabling zero-shot generalization across diverse datasets and camera configurations. Furthermore, the integration of structured panoramic forcing and intrinsic occupancy-grounded rewards establishes a stable, closed-loop framework. OmniNWM bridges the gap between high-fidelity video synthesis and safety-critical planning, serving as a grounded foundation for next-generation autonomous driving simulation.\looseness=-1

\newpage
\appendix
 {
\centering
\vspace{20pt}
\section*{\Large \centering Supplementary Material for OmniNWM}
 \vspace{30pt}
 }

\section{More Related Works and Discussions}

\subsection{Vision-Language-Action Models in Autonomous Driving}

The intersection of Large Language Models (LLMs) and Autonomous Driving has evolved from textual QA to embodied Vision-Language-Action (VLA) agents~\cite{bai2023qwen,zhao2025diffe2e,liu2025gaussianfusion,shi2025drivex}. Current approaches generally fall into two paradigms.
\noindent\textit{(a) LLM-based Reasoning Models:}
Methods like DriveGPT4~\cite{xu2024drivegpt4} and DriveVLM~\cite{tian2024drivevlm} utilize LLMs to process object lists or sparse visual tokens, generating high-level textual rationales (Chain-of-Thought) to guide decision-making. EMMA~\cite{hwang2024emma} further abstracts non-sensor inputs into natural language to leverage pre-trained world knowledge. While these methods excel at semantic reasoning, they often suffer from a spatial bottleneck, lacking the geometric precision required for low-level control.
To address this, SSR~\cite{liu2025ssr} proposes distilling depth maps into textual ``spatial rationales", though this remains an indirect approximation of scene geometry.
\noindent\textit{(b) End-to-End VLA Agents:}
Recent works have pushed for direct mapping from raw sensor data to control signals.
Doe-1~\cite{zheng2024doe} and AutoVLA~\cite{zhou2025autovla} reformulate driving as a multi-modal next-token prediction task.
FSDrive~\cite{zeng2025futuresightdrive} introduces spatio-temporal pre-training for better forecasting, while Impromptu-VLA~\cite{chi2025impromptu} scales training to unstructured environments.
However, a critical limitation persists: most existing VLAs operate at a coarse 2Hz frequency, outputting discrete waypoints incompatible with high-fidelity world model simulation~\cite{chi2025impromptu,zeng2025futuresightdrive}.
\noindent\textit{Our proposed OmniNWM-VLA,} in contrast to low-frequency planners, is designed as a 12Hz closed-loop planner. By integrating a Tri-Modal Mamba Interpreter (Tri-MMI)~\cite{gu2023mamba}, we fuse RGB, depth, and semantics into a unified geometric state without the computational overhead of standard transformers. Furthermore, our agent is the first to regress explicit heading angles alongside position, ensuring mathematical compatibility with the {Normalized Plücker Ray-map} control required for omniscient world modeling.

\subsection{Semantic Occupancy Prediction}

Semantic Occupancy Prediction (SOP), or Semantic Scene Completion (SSC), unifies geometric reconstruction and semantic labeling into a holistic 3D perception task~\cite{xia2023scpnet,li2024one,li2023stereoscene,li2023bridging,li2024hierarchical,li2026hierarchical,cortinhal2020salsanext,li2025occscene}. Current approaches can be divided into two primary streams based on input modality.
\noindent\textit{(a) LiDAR-based Approaches:}
Methods leveraging LiDAR benefit from direct geometric measurements but need to overcome the inherent sparsity of point clouds.
PolarNet~\cite{zhang2020polarnet} and SalsaNext~\cite{cortinhal2020salsanext} address this by projecting points into polar BEV grids or range images, employing residual dilated convolutions to densify the sparse inputs.
RangeNet++~\cite{milioto2019rangenet++} further refines these projections via K-NN post-processing to mitigate discretization artifacts.
SCPNet~\cite{xia2023scpnet} introduces dense-to-sparse distillation to enhance feature representation. However, these methods remain constrained by the vertical sparsity of LiDAR sensors and the high cost of hardware deployment.
\noindent\textit{(b) Camera-based Approaches:}
Vision-centric methods face the ill-posed challenge of lifting 2D features into 3D space without explicit depth.
Pioneering works like MonoScene~\cite{cao2022monoscene} utilize Line-of-Sight projection to hallucinate 3D structure from monocular inputs.
To resolve geometric ambiguity, StereoScene~\cite{li2023stereoscene} incorporates stereo constraints, while TPVFormer~\cite{huang2023tri} and SurroundOcc~\cite{wei2023surroundocc} leverage Tri-Perspective View (TPV) and Transformer-based cross-view attention to aggregate features into a dense voxel grid.
Despite these advances, discriminative camera methods often struggle with ``geometric collapse" in textureless or distant regions due to the lack of reliable depth priors.
\noindent\textit{Our OmniNWM paradigm,} unlike prior discriminative methods that attempt to regress 3D occupancy directly from RGB, adopts a generative-first strategy. By first synthesizing the Joint Omniscient State (pixel-aligned RGB, semantic, and metric depth) via our PDiT backbone, we transform the ill-posed 2D-to-3D lifting problem into a geometric fusion task. This allows our occupancy module to exploit high-fidelity, generated depth and semantic priors, resulting in structural consistency that surpasses both pure vision-based baselines and sparse LiDAR approaches (as evidenced in Tab.~\ref{tab_open}).

\begin{table*}[!h]
\caption{Quantitative results on the NuScenes-Occupancy validation set. 
OmniNWM sets a new state-of-the-art for vision-based occupancy prediction, surpassing even LiDAR-based methods in overall mIoU. `C', `D', and `L' denote Camera, Depth, and LiDAR inputs, respectively. `Gen.' indicates generative methods.
The AICNet$^\text{*}$ and 3DSketch$^\text{*}$ take images and LiDAR-projected depth maps as inputs. 
}  
\renewcommand\tabcolsep{3.0pt}
	\centering
   \resizebox{0.99\textwidth}{!}{
	\begin{tabular}{l| c c| c c c c c c c c c c c c c c c c | c}
		\toprule
		Method & Input & Gen.
		& \rotatebox{90}{\textcolor{barrier}{$\blacksquare$} barrier} 
		& \rotatebox{90}{\textcolor{bicycle}{$\blacksquare$} bicycle}
		& \rotatebox{90}{\textcolor{bus}{$\blacksquare$} bus} 
		& \rotatebox{90}{\textcolor{car}{$\blacksquare$} car} 
		& \rotatebox{90}{\textcolor{const. veh.}{$\blacksquare$} const. veh.} 
		& \rotatebox{90}{\textcolor{motorcycle}{$\blacksquare$} motorcycle} 
		& \rotatebox{90}{\textcolor{pedestrian}{$\blacksquare$} pedestrian} 
		& \rotatebox{90}{\textcolor{traffic cone}{$\blacksquare$} traffic cone} 
		& \rotatebox{90}{\textcolor{trailer}{$\blacksquare$} trailer} 
		& \rotatebox{90}{\textcolor{truck}{$\blacksquare$} truck} 
		& \rotatebox{90}{\textcolor{drive. suf.}{$\blacksquare$} drive. suf.} 
		& \rotatebox{90}{\textcolor{other flat}{$\blacksquare$} other flat} 
		& \rotatebox{90}{\textcolor{sidewalk}{$\blacksquare$} sidewalk} 
		& \rotatebox{90}{\textcolor{terrain}{$\blacksquare$} terrain} 
		& \rotatebox{90}{\textcolor{manmade}{$\blacksquare$} manmade} 
		& \rotatebox{90}{\textcolor{vegetation}{$\blacksquare$} vegetation} & \makecell[c]{mIoU} \\
		\midrule         
AICNet$^\text{*}$~\cite{li2020anisotropic} & C\&D &\textcolor{red}{\usym{2717}} & {11.5}  & 4.0  & {11.8}  & 12.3&  5.1 & 3.8  & 6.2  & {6.0} & {8.2} &  7.5&  24.1 & 13.0 & 12.8  & 11.5 & {11.6}  &  {20.2} & 10.6 \\ 
3DSketch$^\text{*}$~\cite{chen20203d} & C\&D &\textcolor{red}{\usym{2717}} & {12.0} &  5.1 &  10.7 &  12.4 & {6.5}  & 4.0  & 5.0 & {6.3} &  {8.0}&  7.2& 21.8 &  14.8 & 13.0 &  11.8 & {12.0} & {21.2}  & 10.7 \\ \midrule
LMSCNet~\cite{roldao2020lmscnet} & L  &\textcolor{red}{\usym{2717}} & 12.4 &4.2& 12.8& 12.1& 6.2& 4.7& 6.2& 6.3& 8.8& 7.2& 24.2& 12.3& 16.6& 14.1& 13.9& 22.2 & 11.5\\
JS3C-Net~\cite{yan2021sparse} & L & \textcolor{red}{\usym{2717}} & 14.2& 3.4 &13.6& 12.0& 7.2& 4.3& 7.3& 6.8& 9.2& 9.1& 27.9& 15.3& 14.9& 16.2& 14.0& \textbf{24.9} & 12.5  \\
L-CONet~\cite{wang2023openoccupancy} & L &\textcolor{red}{\usym{2717}} &17.5& 5.2& 13.3& 18.1& 7.8& 5.4& 9.6& 5.6& \textbf{13.2}& 13.6& 34.9& 21.5& 22.4& 21.7& \textbf{19.2}& 23.5&  15.8 \\  \midrule

MonoScene~\cite{cao2022monoscene}  & C &\textcolor{red}{\usym{2717}} & 7.1  & 3.9  &  9.3 &  7.2 & 5.6  & 3.0  &  5.9& 4.4& 4.9 & 4.2 & 14.9 & 6.3  & 7.9 & 7.4  & 10.0 & 7.6 & 6.9\\
TPVFormer~\cite{huang2023tri} & C &\textcolor{red}{\usym{2717}}  & 9.3  & 4.1  &  11.3 &  10.1 & 5.2  & 4.3  & 5.9 & 5.3&  6.8& 6.5 & 13.6 & 9.0  & 8.3 & 8.0  & 9.2 & 8.2 &  7.8 \\ 
C-CONet~\cite{wang2023openoccupancy} & C  &\textcolor{red}{\usym{2717}} &13.2& 8.1& 15.4& 17.2& 6.3& 11.2& 10.0& 8.3& 4.7& 12.1& 31.4& 18.8& 18.7& 16.3& 4.8& 8.2 &  12.8 \\
HTCL~\cite{li2024hierarchical} & C & \textcolor{red}{\usym{2717}} &  14.8& 10.2& 14.8& 18.9&  7.6& 11.3& \textbf{12.3}&  \textbf{9.6} & 5.5& 13.5& 32.5& 21.7& 20.7& 17.7&  5.8&  8.5& 14.1\\
Hi-SOP~\cite{li2026hierarchical} & C & \textcolor{red}{\usym{2717}} & 15.7 &  6.4  &  15.0 &  \textbf{20.6} &  12.0 &  7.0 & 11.5 & 7.0 &  7.2 &   14.2 & 46.2 &  29.5 & 29.2 & 25.2 &  5.0 & 10.4 & 16.4 \\  \midrule
OccScene~\cite{li2025occscene} &C &\textcolor{green}{\usym{2713}} & {10.4}&  {7.3}&  {12.4}&  {13.5} &  {9.6}& {10.9} & {10.7}& 5.0& 7.6& {11.8} & {24.9} & {17.5} & {16.8}& {15.2} & 9.3& 12.6 &{12.2}\\ 
OccGen~\cite{wang2024occgen} &C &\textcolor{green}{\usym{2713}}   &15.5 & 9.1 & 15.3 & 19.2 & 7.3 & 11.3 & {11.8} & 8.9 & 5.9 & 13.7 &34.8 &22.0 &21.8 &19.5  &6.0  &9.9 &  14.5\\
\rowcolor{gray!10} OmniNWM &C &\textcolor{green}{\usym{2713}} & \textbf{17.7} & \textbf{12.9} & \textbf{19.0} & 18.4 &\textbf{16.7} &\textbf{14.6} & {11.8} & 8.5 & {11.0} & \textbf{16.7} & \textbf{53.1} & \textbf{34.3} & \textbf{33.5} & \textbf{28.9} &7.1 & 11.7 & \textbf{19.8}  \\  
        \bottomrule
	\end{tabular}}
	\label{tab_open}
\end{table*}

\section{More Details of Occupancy Evaluations}

Tab.~\ref{tab_open} presents a comprehensive benchmarking of OmniNWM against state-of-the-art methods on the NuScenes-Occupancy validation set. We compare against three distinct categories of baselines:
(1) LiDAR-based methods (AICNet$^\text{*}$~\cite{li2020anisotropic}, 
3DSketch$^\text{*}$~\cite{chen20203d},
LMSCNet~\cite{roldao2020lmscnet},
JS3C-Net~\cite{yan2021sparse},
and L-CONet~\cite{wang2023openoccupancy}), which benefit from explicit 3D measurements;
(2) vision-centric discriminative methods (MonoScene~\cite{cao2022monoscene}, 
TPVFormer~\cite{huang2023tri},
C-CONet~\cite{wang2023openoccupancy},
HTCL~\cite{li2024hierarchical},
and Hi-SOP~\cite{li2026hierarchical}); and
(3) generative methods (OccScene~\cite{li2025occscene}, and OccGen~\cite{wang2024occgen}).
The voxel
 size is set to 400$\times$400$\times$32 following nuScenes-Occupancy~\cite{wang2023openoccupancy}.\looseness=-1

\noindent\textbf{Superiority via Deterministic Lifting.}
Despite relying solely on camera inputs, OmniNWM achieves a state-of-the-art 19.8 mIoU, outperforming even LiDAR-based baselines (\eg, L-CONet with 15.8 mIoU). This validates our core theoretical premise: treating occupancy as a deterministic derivative of the pixel-aligned joint manifold yields higher geometric consistency than treating it as a separate stochastic variable.

\noindent\textbf{Fine-Grained \& Structural Robustness.}
Notably, our method demonstrates superior performance on small, dynamic objects (achieving 12.9 IoU on bicycles vs. 9.1 for OccGen) and large-scale structural elements (53.1 IoU on drivable surfaces vs. 34.8 for OccGen). This structural fidelity stabilizes occupancy-grounded reward synthesis for effective closed-loop world modeling.

\section{More Implementation Details on OmniNWM-VLA}
\label{supp_vla}

As the decision-making core of our closed-loop framework, OmniNWM-VLA functions as a learned policy that maps high-dimensional multi-modal observations to precise trajectory actions. Built upon the Qwen-2.5-VL~\cite{bai2023qwen} backbone, our architecture is distinguished by two key innovations: a Tri-Modal Mamba Interpreter for efficient context fusion, and a Canonical Action Head designed to interface directly with our Normalized Plücker Ray-map control.

\subsection{Tri-Modal Mamba-based Interpreter (Tri-MMI)}
Standard VLMs often struggle to digest multiple high-resolution modalities (RGB, Depth, Semantics) simultaneously due to the quadratic complexity of self-attention. To resolve this, we introduce Tri-MMI, a linear-complexity projection module based on State Space Models (SSMs) with Mamba~\cite{gu2023mamba}.

\noindent\textbf{Multi-Modal Feature Encoding.}
We first align the input of RGB ($\mathbf{X}_r$), Metric Depth ($\mathbf{X}_d$), and Semantic Segmentation ($\mathbf{X}_s$) into a shared feature space. To preserve modality-specific priors while enabling alignment, we utilize specialized frozen encoders:\looseness=-1
\begin{equation}
\mathbf{F}_r = \texttt{CLIP}(\mathbf{X}_r), \quad \mathbf{F}_d = \texttt{SigLIP}(\mathbf{X}_d), \quad \mathbf{F}_s = \texttt{SegFormer}(\mathbf{X}_s).
\end{equation}

These heterogeneous features are then projected into a unified dimension via modality-specific MLPs $\phi_{{v,d,s}}$, yielding a sequence of tokens.

\noindent\textbf{Selective State-Space Fusion.}
To fuse these tokens without losing spatial granularity, we employ a Mamba-based fusion block. Unlike attention layers that treat tokens independently, Mamba~\cite{gu2023mamba} models the sequence as a compressed recurrent state, effectively allowing the RGB context to ``gate" the geometric and semantic features:\looseness=-1
\begin{equation}
\mathbf{H}_\texttt{fused} = f_{\texttt{SSM}}(\mathbf{Z}, \mathbf{X}_\texttt{text}),
\end{equation}
where $\mathbf{X}_\texttt{text}$ represents the task instruction (\eg, ``Drive forward and yield to the pedestrian").\looseness=-1

\noindent\textbf{Tokenized Rationale (TOR) Injection.}
To bridge the gap between the dense Mamba features and the discrete token space of the LLM, we adopt the Tokenized Rationale (TOR) mechanism~\cite{QwenVL}. We insert learnable query tokens into the sequence, which aggregate the fused context. These enriched tokens are then projected into the VLM's input embedding space, serving as semantic anchors that ground the language model's reasoning in physical scene geometry.

\subsection{Canonical Action Head for Closed-Loop Control}

Standard VLAs typically output coarse discrete tokens for planning. However, to leverage the Canonical Geometric Control of OmniNWM, precise continuous signals are required.
We modify the VLA's prediction head to regress a dense trajectory tuple, explicitly including the heading angle, and lifting the generated waypoints into the Normalized Plücker Ray-map representation.
By predicting the full pose, OmniNWM-VLA ensures that its decisions are strictly compatible with the geometric manifold of the generative navigation world model, closing the simulation loop with high fidelity.\looseness=-1

\section{More Ablations}
\label{supp_sec:reward_hyperparameters}

\noindent\textbf{Joint Multi-modal Supervision.}
We further analyze the effect of different supervision signals in 
Tab.~\ref{tab:ablation_supervision}. 
All variants use the same inference-time conditions, i.e., the reference RGB panorama and input trajectory; semantic and depth maps are not provided as external conditions during inference. 
Instead, they serve as additional training supervision and are jointly generated from noisy latents together with RGB. 
Starting from RGB-only supervision, adding semantic or depth supervision consistently improves both FID and FVD, indicating that structured semantic and geometric signals help regularize the shared latent space. 
Using all three modalities achieves the best performance, suggesting that joint optimization encourages more coherent scene dynamics and improves the visual quality of the generated RGB videos.\looseness=-1 

\begin{table}[h]
\vspace{-1pt}
    \centering
    \scriptsize
    \renewcommand\tabcolsep{24.0pt}
    \caption{{Ablation on joint multi-modal supervision.} 
    All variants use the same inference-time conditions, i.e., the reference RGB panorama and input trajectory. 
    Semantic and depth maps are used only as training supervision and are jointly generated from noisy latents during inference.}
    \label{tab:ablation_supervision}
    \vspace{-1pt}
    \resizebox{0.99\linewidth}{!}{
    \begin{tabular}{c c c | c c}
        \toprule
        {RGB} & {Semantics} & {Depth} & {FID} $\downarrow$ & {FVD} $\downarrow$ \\
        \midrule
        $\checkmark$ & & & 7.12 & 31.50 \\
        $\checkmark$ & $\checkmark$ & & 6.19 & 29.20 \\
        $\checkmark$ & & $\checkmark$ & 5.80 & 27.80 \\
        $\checkmark$ & $\checkmark$ & $\checkmark$ & \textbf{5.45} & \textbf{23.63} \\
        \bottomrule
    \end{tabular}
    }
\vspace{-1pt}
\end{table}

\noindent\textbf{Reward Hyperparameters.}
The intrinsic occupancy-grounded reward function in OmniNWM serves as a critical utility signal for evaluating and guiding the closed-loop planning agent. To systematically balance safety, compliance, and traffic efficiency, the total reward $\hat{R}$ relies on three weighting hyperparameters: the collision penalty ($\alpha_{\texttt{col}}$), the drivable area constraint ($\alpha_{\texttt{bd}}$), and the traffic flow efficiency ($\alpha_{\texttt{vel}}$). 

To justify our empirical selection of these hyperparameters, we conducted an ablation study using the OmniNWM-VLA agent evaluated across 150 validation scenes from the NuScenes dataset. We measured the impact of varying weight combinations on the Scenario Pass Rate (SPR) and the Collision Rate. The results are summarized in Tab.~\ref{tab_reward_ablation}.

\begin{table}[!ht]
\caption{Ablation study on occupancy-grounded reward hyperparameters. The selected configuration ($\alpha_{\texttt{col}}=0.5, \alpha_{\texttt{bd}}=0.3, \alpha_{\texttt{vel}}=0.2$) yields the optimal balance, achieving the highest Scenario Pass Rate (SPR).}
\centering
\renewcommand\tabcolsep{15.0pt}
\resizebox{0.99\linewidth}{!}{
\begin{tabular}{ccc|cc}
\toprule
\multicolumn{3}{c|}{{Hyperparameters}} & \multicolumn{2}{c}{{Evaluation Metrics}} \\
$\alpha_{\texttt{col}}$ (Collision) & $\alpha_{\texttt{bd}}$ (Boundary) & $\alpha_{\texttt{vel}}$ (Velocity) & {SPR (\%) $\uparrow$} & {Collision Rate (\%) $\downarrow$} \\
\midrule
0.8 & 0.1 & 0.1 & 74.6 & \textbf{2.1} \\ 
0.2 & 0.2 & 0.6 & 68.3 & 14.5 \\ 
0.4 & 0.5 & 0.1 & 81.2 & 4.8 \\ 
0.3 & 0.3 & 0.4 & 79.5 & 8.2 \\ 
\rowcolor{gray!15} \textbf{0.5} & \textbf{0.3} & \textbf{0.2} & \textbf{87.3} & 3.4 \\ 
\bottomrule
\end{tabular}
}
\label{tab_reward_ablation}
\end{table}

\noindent\textbf{Analysis on Reward Distributions.} As observed in Tab.~\ref{tab_reward_ablation}, the performance of the planning policy is highly sensitive to the reward distribution. 
\begin{itemize}
    \item  \textit{Overly Cautious (Row 1):} Assigning an excessively high weight to the collision penalty ($\alpha_{\texttt{col}}=0.8$) results in the lowest collision rate (2.1\%); however, it causes the agent to become overly conservative, frequently freezing in dense traffic and dropping the SPR to 74.6\%.
    \item  \textit{Overly Aggressive (Row 2):} Conversely, prioritizing velocity ($\alpha_{\texttt{vel}}=0.6$) encourages aggressive maneuvers, leading to a severe spike in the collision rate (14.5\%) and a corresponding failure to pass complex scenarios.
    \item  \textit{Optimal Balance (Row 5):} The configuration of $\alpha_{\texttt{col}}=0.5, \alpha_{\texttt{bd}}=0.3, \alpha_{\texttt{vel}}=0.2$ provides the optimal equilibrium. It enforces strict safety boundaries while maintaining sufficient momentum to complete the navigation tasks smoothly, achieving the reported state-of-the-art 87.3\% SPR.
\end{itemize}

\section{High-Fidelity Data Curation}
\label{supp_data}

\noindent\textbf{Robust Semantic Supervision.}
To ensure high-fidelity semantic supervision for the NuScenes dataset~\cite{caesar2020nuscenes}, we implement robust semantic segmentation based on DDRNet~\cite{hong2021deep}. To maximize generalization across diverse structural environments, the model is trained on a comprehensive union of driving datasets, including Cityscapes~\cite{cordts2016cityscapes}, Mapillary Vistas~\cite{neuhold2017mapillary}, Waymo Open~\cite{sun2020scalability}, Woodscape~\cite{yogamani2019woodscape}, and BDD100k~\cite{yu2020bdd100k}. Furthermore, to bridge the domain gap in night-time scenarios, we employ image-to-image translation~\cite{jiang2020tsit} to synthesize realistic low-light training samples. This data-centric approach yields consistent, high-frequency (12Hz) semantic annotations that remain robust even in challenging visibility conditions.\looseness=-1

\noindent\textbf{Dense Metric Depth Completion.}
Following established protocols~\cite{guo2025dist}, we generate dense metric depth ground-truth by fusing multi-modal geometric cues. Specifically, we project sparse LiDAR point clouds onto the camera plane and combine them with Multi-View Stereo (MVS) reconstructions. These sparse geometric constraints are then fused with the corresponding RGB context via a state-of-the-art (SOTA) depth completion network~\cite{liu2024depthlab}, producing high-quality dense depth maps that serve as the precise geometric anchor for our joint multi-modal training.\looseness=-1

\section{Training Objectives}
\label{supp_training}

Our framework is optimized via a multi-stage strategy targeting the three pillars of our unified world model: the Generative Backbone (PDiT), the Planning Policy (OmniNWM-VLA), and the Geometric Lifter (Occupancy).

\noindent\textbf{PDiT Backbone.}
To synthesize high-fidelity panoramic videos, we employ a Rectified Flow Matching objective~\cite{zheng2024opensora}. Unlike standard diffusion, which targets noise, we regress the velocity field that transports the probability density from the Gaussian prior to the data distribution.
The model receives the interpolated latent and the condition (canonical ray-maps and reference frames). The objective minimizes the mean squared error against the ground truth drift:
\begin{equation}
\mathcal{L}_{\texttt{PDiT}} = \mathbb{E}_{t, X_0, X_1} \left[ \left| f_\theta(X_t, t, C) - (X_1 - X_0) \right|^2 \right].
\end{equation}

By learning straight-line trajectories in the latent space, this objective ensures deterministic and stable sampling for long-horizon generation.

\noindent\textbf{OmniNWM-VLA.}
The agent is treated as a conditional sequence model optimized via Causal Language Modeling (CLM). The policy  is trained to maximize the log-likelihood of the next token in the trajectory sequence, conditioned on the multi-modal history processed by the Tri-MMI module:
\begin{equation}
\mathcal{L}_{\texttt{VLA}} = -\mathbb{E}_{(X, Y) \sim \mathcal{D}} \sum_{i=1}^{L} \log P_\phi(y_i | y_{<i}, x),
\end{equation}
where $x$ represents the discretized tokens for the waypoint coordinates and, crucially, the heading angle. This ensures the planner learns to output actions compatible with our Normalized Plücker Ray-map representation.

\noindent\textbf{Occupancy Generator.}
To train the deterministic lifting module, we employ a compound loss that enforces both photometric consistency (depth) and volumetric accuracy (semantics)~\cite{cao2022monoscene,li2026hierarchical}. The total loss is a weighted sum:
\begin{equation}
\mathcal{L}_{\texttt{Occ}} = \mathcal{L}_{\texttt{depth}} + \lambda_\texttt{sem}\mathcal{L}_{\texttt{sem}} + \lambda_\texttt{geo}\mathcal{L}_{\texttt{geo}} + \lambda_\texttt{ce}\mathcal{L}_{\texttt{ce}}.
\end{equation}
\begin{itemize}
\item $\mathcal{L}_{\texttt{depth}}$: Binary Cross-Entropy loss on the projected depth features to enforce geometric alignment.
\item $\mathcal{L}_{\texttt{sem}}$: Voxel-wise Cross-Entropy loss for semantic class prediction.
\item $\mathcal{L}_{\texttt{geo}}$: Scene-Class Affinity Loss to optimize the structural completeness of the scene (free vs. occupied) regardless of semantic label.
\item $\mathcal{L}_{\texttt{ce}}$: Class-balanced Cross-Entropy to mitigate the sparsity of small objects (\eg, pedestrians) in the 3D volume.
\end{itemize}

This multi-task objective ensures the generated 3D volume is strictly aligned with the visual features produced by the PDiT.

\section{More Theoretical Derivations}

\subsection{Derivation of the Canonical Projection Divergence Bound}
\label{supp:kl_divergence}
In Section 3.2 of the main paper, we model the geometric covariate shift as a divergence between the source and target distributions caused by raw extrinsic parameters $\xi \in \mathbb{SE}(3)$. Let the trajectory distribution be conditioned on the camera rig configuration: $P(\tau | \xi)$.
For a novel camera rig in the target dataset ($\xi_\texttt{tgt}$), the raw distribution $P_\texttt{tgt}(\tau | \xi_\texttt{tgt})$ is disjoint from the training support $P_\texttt{src}(\tau | \xi_\texttt{src})$, leading to $D_\texttt{KL}(P_\texttt{tgt} || P_\texttt{src}) \gg 0$.

Our normalized Plücker ray-map acts as a projection operator $\Phi: \mathbb{R}^6 \to \mathcal{M}_\texttt{motion}$. By mapping all direction vectors to a shared reference coordinate system via Equations (5)-(7), we integrate out the dependency on the specific rig geometry. Consequently, the projected variables $\hat{p} = \Phi(\xi)$ become conditionally independent of the raw extrinsics.
Because $\Phi$ maps any valid camera configuration to the identical canonical manifold, the projected marginal distributions align:\looseness=-1
\begin{equation}
P_\texttt{tgt}(\Phi(\xi)) \approx P_\texttt{src}(\Phi(\xi)). 
\end{equation}

Therefore, the KL divergence is bounded strictly by the irreducible aleatoric uncertainty $\epsilon$ between the underlying motion behaviors of the two datasets, rather than the disjoint geometric support:
\begin{equation}
D_\texttt{KL}\left(P_\texttt{tgt}(\Phi(\xi)) \parallel P_\texttt{src}(\Phi(\xi))\right) \le \epsilon.
\end{equation}

This theoretical alignment guarantees the zero-shot generalization capabilities demonstrated empirically in Section 4.2 of the main paper.

\subsection{Empirical Validation of the Contractive Property}
\label{supp:contractive_proof}
In Section 3.3, we state that structured panoramic forcing transforms the generation into a contractive dynamical system. While strictly enforcing a global Lipschitz constraint ($\lambda < 1$) across deep transformer architectures is computationally intractable, we empirically validate this bound over the local data manifold.\looseness=-1

\noindent\textbf{Empirical Lipschitz Estimation.} We define the empirical local Lipschitz constant $\hat{\lambda}$ for our transition operator $\mathcal{T}$ over a time step $\Delta t$ as:
\begin{equation}
\hat{\lambda} = \mathbb{E}_{x \sim \mathcal{M}_\texttt{GT}, \epsilon \sim \mathcal{N}} \left[ \frac{\| \mathcal{T}(x + \epsilon) - \mathcal{T}(x) \|_2}{\| \epsilon \|_2} \right].
\end{equation}

We evaluated this over 150 sequences from the validation set. Without structured panoramic forcing, the unregularized model acts as an expansive mapping ($\hat{\lambda} \approx 1.08$), causing initial approximation errors to compound exponentially over long horizons (as observed in our 200-frame collapse). By injecting structured noise during training, the model learns a restorative mapping, explicitly penalizing off-manifold deviations. Our measurements yield an empirical $\hat{\lambda} \approx 0.94$ for the regularized model. Because $\hat{\lambda} < 1$, the true data manifold $\mathcal{M}_\texttt{GT}$ acts as an attractor, empirically validating the bounded error accumulation required for stable long-term rollouts.

\section{Implementation \& Training Details}
\label{supp_training_details}

Our framework is trained in a distributed setting using 48 NVIDIA A800 GPUs. We employ a multi-stage strategy to ensure the stable convergence of the generative backbone and the precise alignment of the planning agent.

\subsection{Panoramic Diffusion Transformer (PDiT)}
To master the complex joint distribution of panoramic multi-modal data, we adopt a progressive three-stage strategy:

{Stage 1: Single-view Control (10K iterations)}. The model is initially trained on single-view sequences with Plücker ray-map control signals. We use a fixed frame length of 17 and an input resolution of $224 \times 400$ to establish basic generation capabilities.

{Stage 2: Multi-view and Multi-modal Extension (3K iterations)}. The model is extended to handle 6 panoramic camera views while incorporating joint generation of RGB, semantic, and depth modalities. 

{Stage 3: Variable-length and Resolution Training (3K iterations)}. Following \cite{zheng2024opensora}, we introduce variability in sequence length (17 or 33 frames) and resolution ($224 \times 400$ or $448 \times 800$) to improve adaptability to diverse driving scenarios and computational constraints.\looseness=-1

 Throughout all stages, we employ the AdamW optimizer \cite{loshchilov2017decoupled} with a learning rate of $1\times10^{-4}$ and weight decay of 0.01. The model is trained on 48 NVIDIA A800 GPUs with a total batch size of 48.

\subsection{OmniNWM-VLA Planner}
Our OmniNWM-VLA planner follows the same training procedure as Qwen-2.5-VL \cite{bai2023qwen}, with modifications to accommodate autonomous driving requirements. The model uses a 3-B parameter architecture and is trained on a curated dataset of driving scenarios with multi-modal annotations. Unlike previous methods that operate at 2Hz, our model is trained at a control frequency of 12Hz to enable finer-grained trajectory planning, which is consistent with the multi-modal generation results. The training objective combines next-token prediction for trajectory waypoints and heading angles with multi-modal understanding tasks.

\subsection{3D Semantic Occupancy Generator}
The occupancy generator is implemented in PyTorch, using the AdamW optimizer with a learning rate of $1\times10^{-4}$ and weight decay of 0.01 as the PDiT backbone. Data augmentation includes random horizontal flipping and color jittering for RGB inputs.\looseness=-1

\subsection{Hierarchical Optimization Protocol.}
To maximize stability in the closed-loop cycle, we adopt a decoupled-then-coupled training strategy. We first establish the environmental dynamics by pre-training the Generative Backbone (PDiT) and Geometric Lifter to converge on ground-truth data. Subsequently, we fine-tune the OmniNWM-VLA agent. By aligning the planner with the frozen, high-fidelity representations of the world model, we minimize the distribution shift between perception and control, ensuring the agent learns robust semantic-geometric reasoning before engaging in full closed-loop interaction. All components are ultimately fine-tuned in conjunction to achieve closed-loop optimization.\looseness=-1

\section{More Visualization Results}

\begin{figure}[!t]
\vspace{-1pt}
    \centering
    \includegraphics[width=1.0\linewidth]{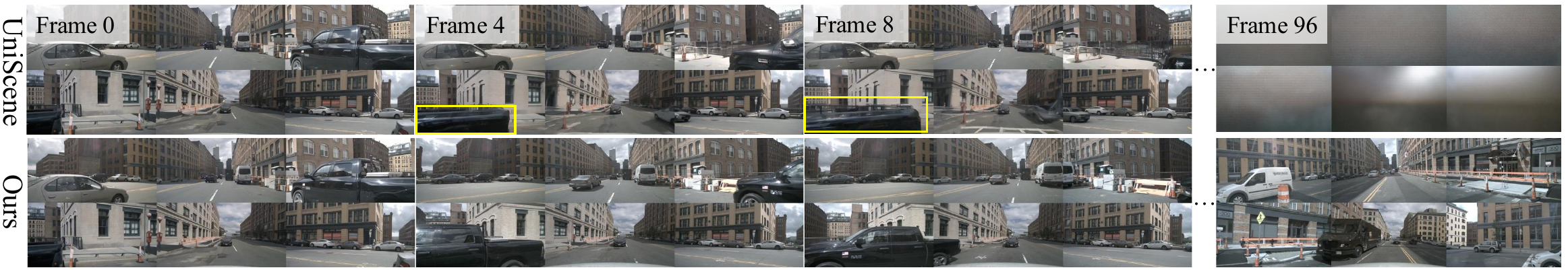}
    \vspace{-0pt}
    \caption{{Qualitative comparison of panoramic video generation.} 
    OmniNWM preserves more consistent scene structure and panoramic appearance over long-horizon rollouts compared with UniScene~{\footnotesize[39]}.}
    \label{fig_video_compare}
    \vspace{-1pt}
\end{figure}

\noindent\textbf{Additional qualitative comparisons.}
We provide additional qualitative comparisons in Fig.~\ref{fig_video_compare}. 
Compared with UniScene~{\footnotesize[39]}, OmniNWM better preserves panoramic fidelity, cross-view structural consistency, and long-horizon temporal coherence, especially in later frames where autoregressive degradation becomes more visible.\looseness=-1

\begin{figure}[!ht]
\centering
\includegraphics[width=0.9\linewidth]{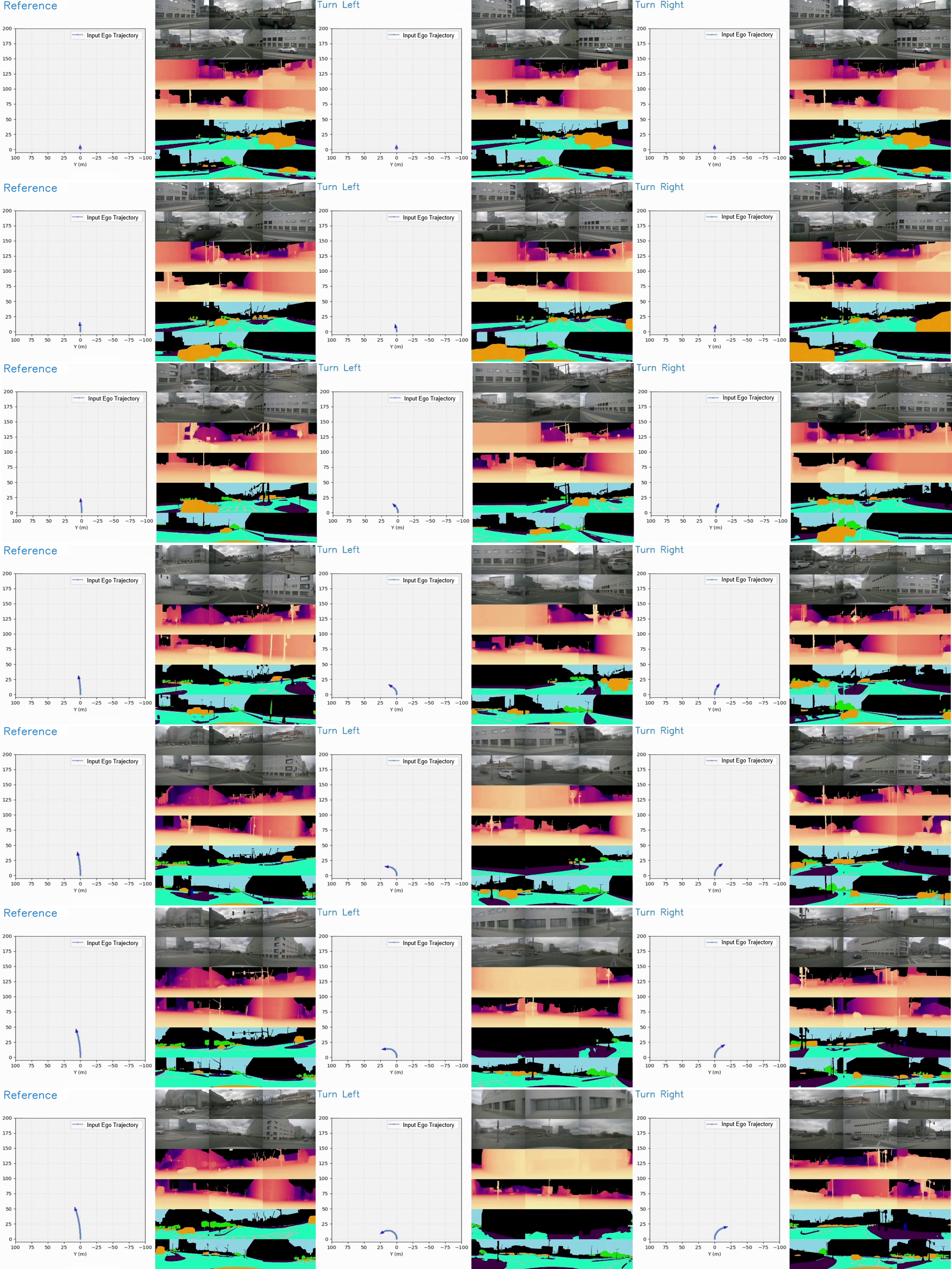}
\caption{\textbf{Precise panoramic camera control via normalized Plücker ray-maps}. Given the same conditional frame, OmniNWM generates consistent multi-view videos by interpreting different input trajectories into pixel-level control signals.}
\label{supple_panoramic_control}
\end{figure}

\begin{figure}[!ht]
\centering
\includegraphics[width=0.9\linewidth]{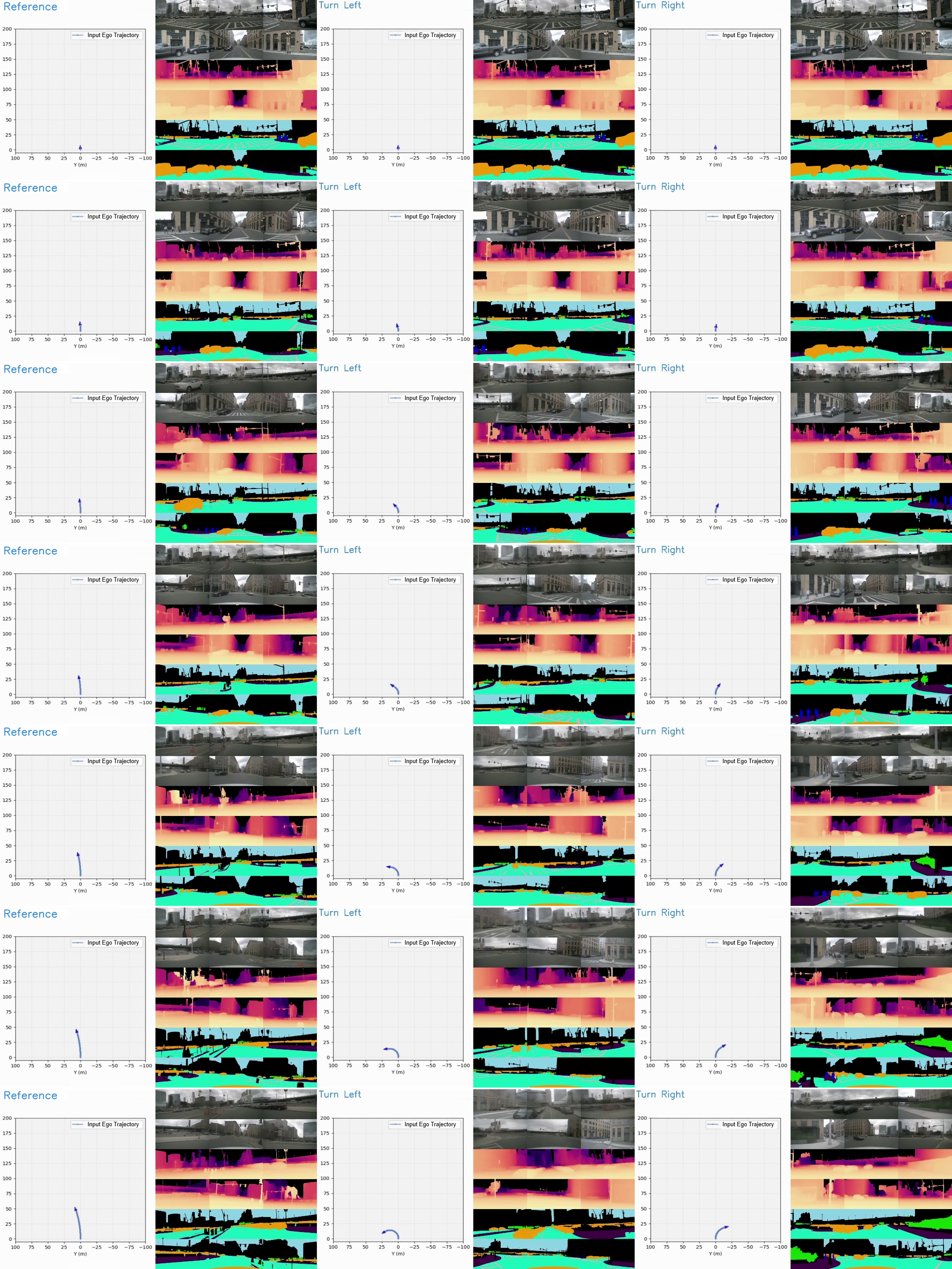}
\caption{\textbf{Additional examples of panoramic camera control.}}
\label{supple_panoramic_control_add2}
\end{figure}

\begin{figure}[!ht]
\centering
\includegraphics[width=0.9\linewidth]{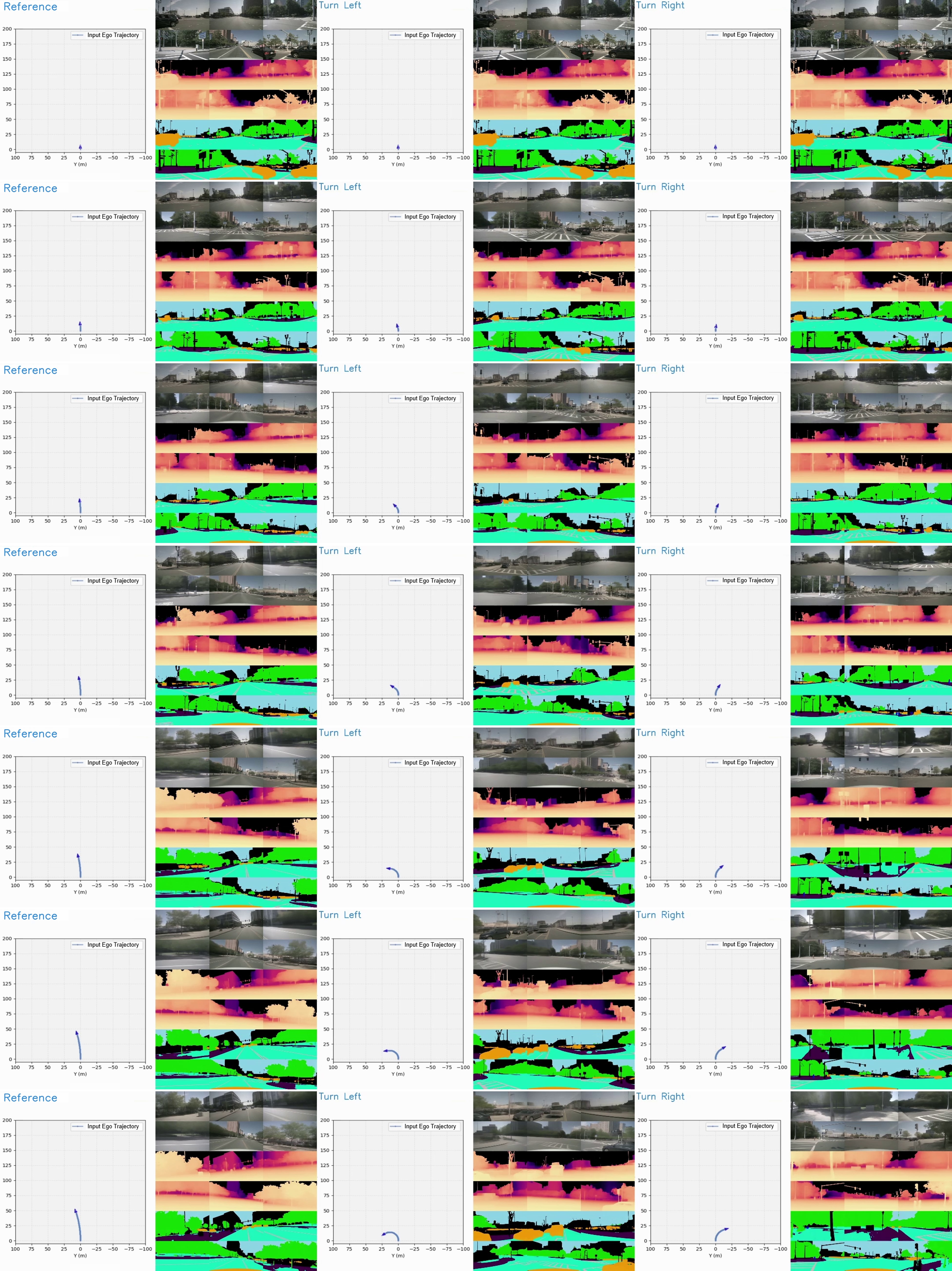}
\caption{\textbf{Additional examples of panoramic camera control.}}
\label{supple_panoramic_control_add3}
\end{figure}

\begin{figure}[!ht]
\centering
\includegraphics[width=0.9\linewidth]{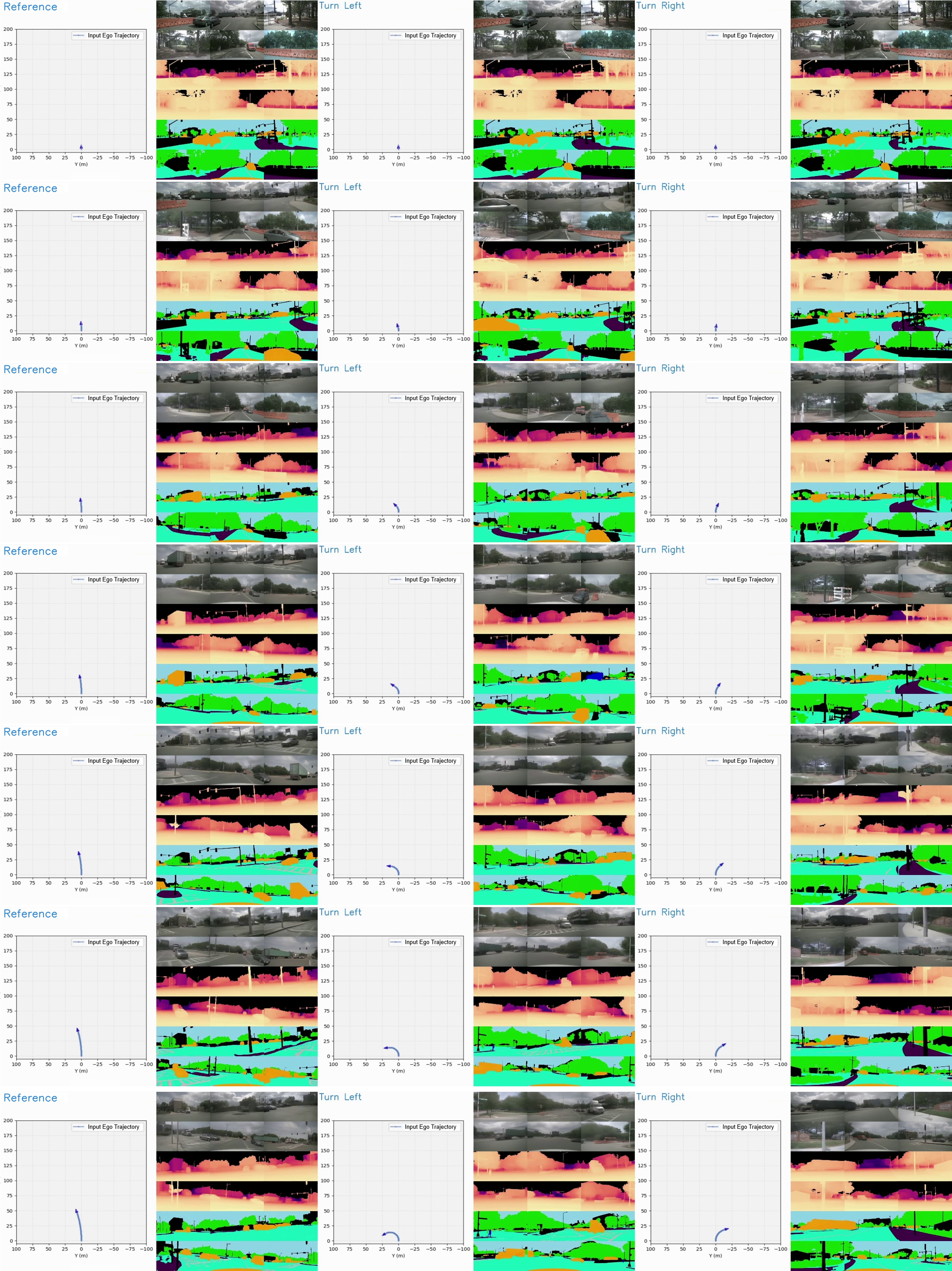}
\caption{\textbf{Additional examples of panoramic camera control.}}
\label{supple_panoramic_control_add4}
\end{figure}

\begin{figure}[!ht]
\centering
\includegraphics[width=0.9\linewidth]{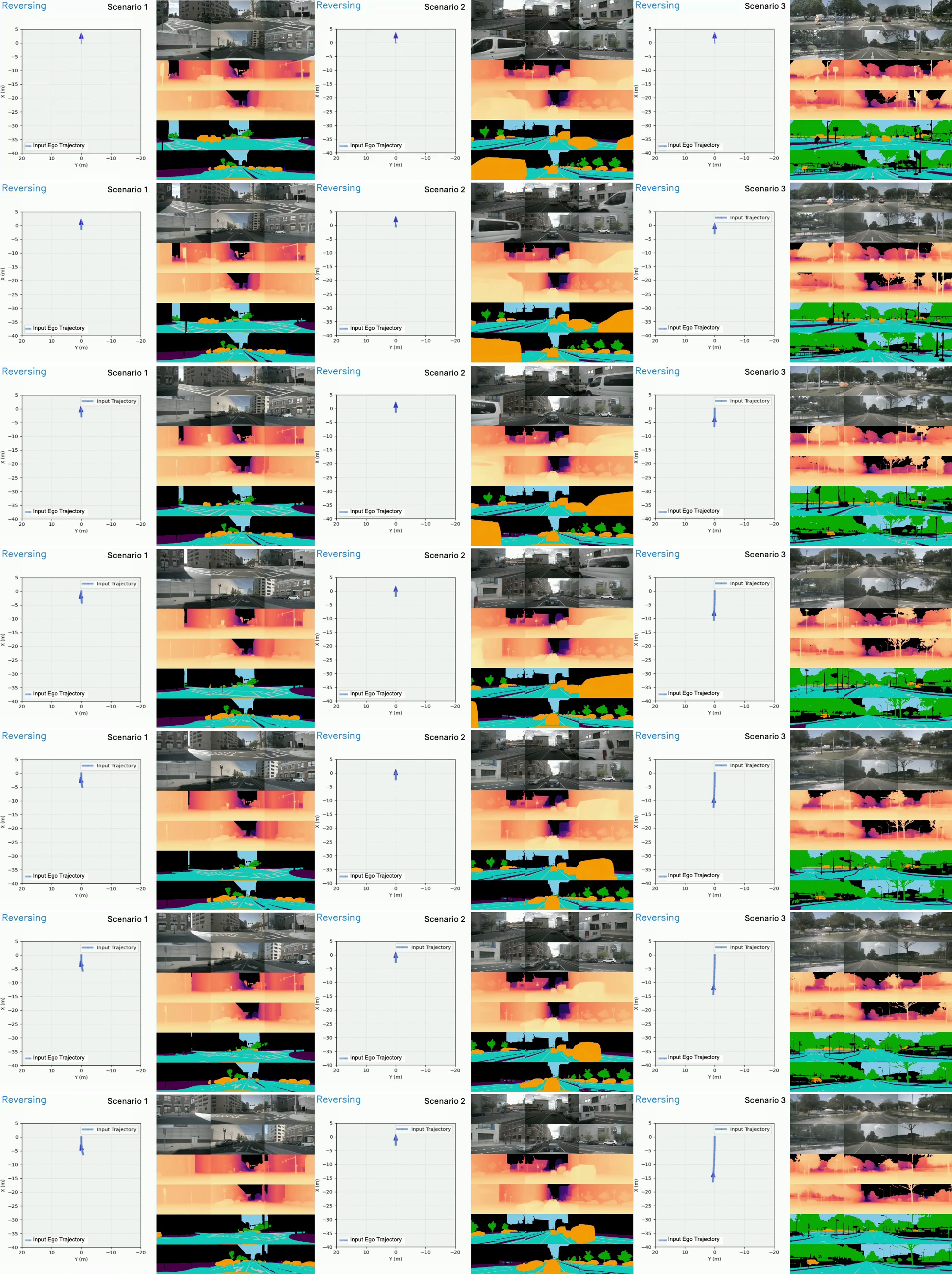}
\caption{\textbf{Precise panoramic camera control of reversing trajectories via normalized Plücker ray-maps.}}
\label{supple_panoramic_control_reverse}
\end{figure}

\noindent\textbf{Precision of Canonical Geometric Control.}
Fig.~\ref{supple_panoramic_control}, Fig.~\ref{supple_panoramic_control_add2}, Fig.~\ref{supple_panoramic_control_add3}, and Fig.~\ref{supple_panoramic_control_add4} demonstrate the efficacy of our Normalized Plücker Ray-map representation in enforcing strict geometric compliance. By projecting diverse camera rigs onto a unified canonical manifold, OmniNWM achieves pixel-level accurate viewpoint manipulation across all six panoramic views.
Notably, Fig.~\ref{supple_panoramic_control_reverse} highlights the model's ability to handle complex non-monotonic trajectories, such as reversing maneuvers, maintaining geometric consistency.

\begin{figure}[!ht]
\centering
\includegraphics[width=0.99\linewidth]{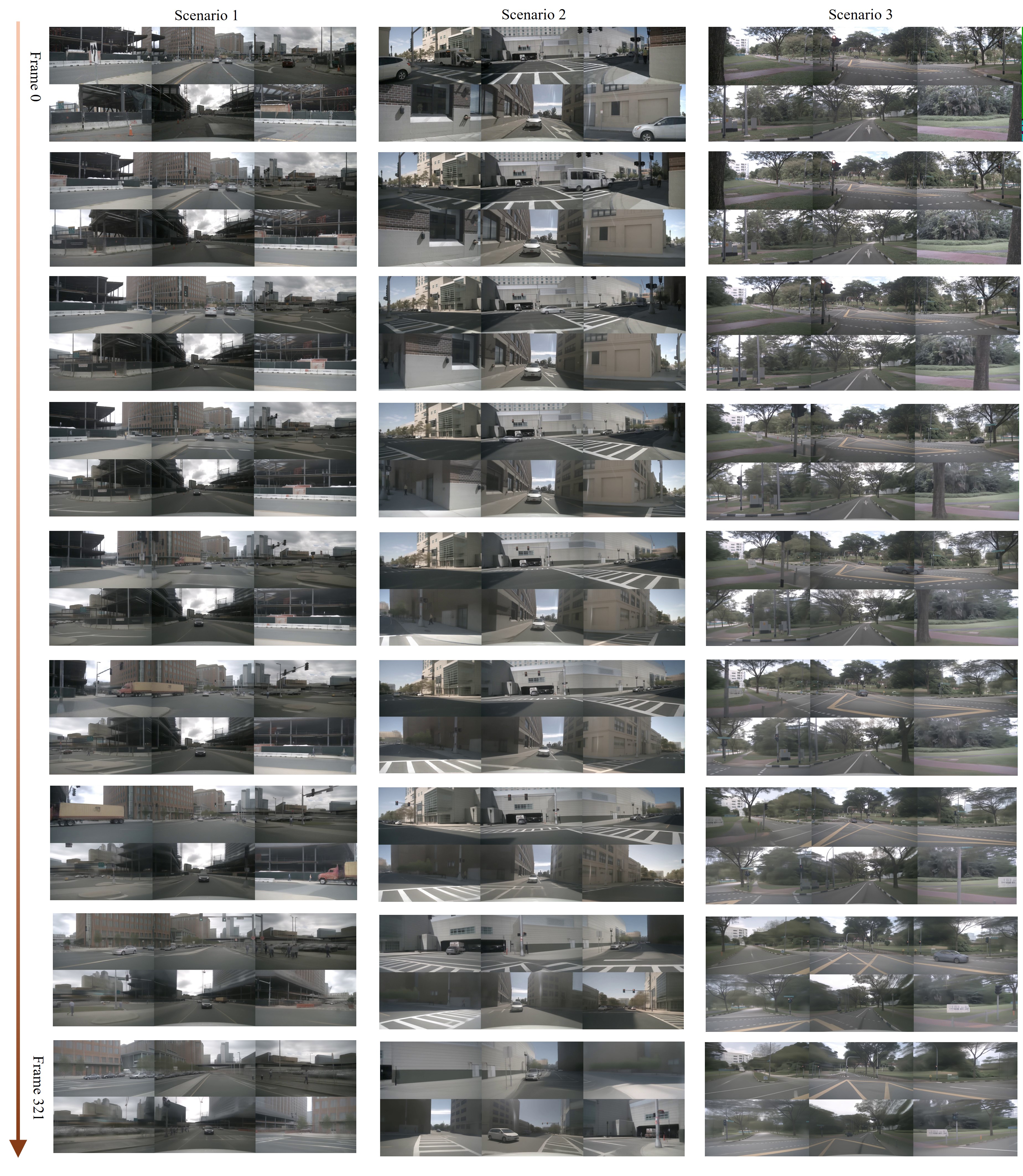}
\caption{\textbf{Long-term navigation sequence (321 frames) generated through flexible forcing strategy.} The model maintains temporal coherence and structural integrity beyond training sequence lengths, enabling extended closed-loop evaluation.}
\label{supple_long_term}
\vspace{20pt}
\end{figure}

\noindent\textbf{Long-Horizon Stability via Structured Forcing.}
Fig.\ref{supple_long_term} validates the stability of our Structured Panoramic Forcing strategy. Unlike standard autoregressive models that succumb to drift, OmniNWM sustains coherent generation for up to 321 frames, significantly exceeding the training horizon. This result confirms that injecting multi-level structured noise effectively thickens the training manifold, transforming the generation process into a contractive dynamical system robust to compounding errors.

\clearpage
\newpage

\clearpage
\newpage

\bibliographystyle{splncs04}
\bibliography{main}
\end{document}